\documentclass[10pt,twocolumn,letterpaper]{article}

\usepackage{cvpr}
\usepackage{times}
\usepackage{epsfig}
\usepackage{graphicx}
\usepackage{amsmath}
\usepackage{amssymb}
\usepackage{cite}
\usepackage{algorithm}
\usepackage{algpseudocode}
\usepackage{enumerate}
\usepackage{dsfont}
\usepackage{subcaption}
\usepackage{float}
\usepackage{graphicx}
\usepackage[table]{xcolor}
\usepackage{textcomp}
\usepackage{xcolor}

\cvprfinalcopy 

\def\cvprPaperID{****} 

\begin{document}

\title{On the Transferability of Winning Tickets in Non-Natural Image Datasets}

\author{Matthia Sabatelli
\\ Universit\'e de Li\`ege\\
{\tt\small m.sabatelli@uliege.be}
\and
Mike Kestemont\\
Universiteit Antwerpen\\
{\tt\small mike.kestemont@uantwerpen.be}
\and
Pierre Geurts\\
Universit\'e de Li\`ege\\
{\tt\small p.geurts@uliege.be}}

\maketitle

\begin{abstract}
We study the generalization properties of pruned models that are the winners of the lottery ticket hypothesis on photorealistic datasets. We analyse their potential under conditions in which training data is scarce and comes from a not-photorealistic domain. More specifically, we investigate whether pruned models that are found on the popular CIFAR-10/100 and Fashion-MNIST datasets, generalize to seven different datasets coming from the fields of digital pathology and digital heritage. Our results show that there are significant benefits in training sparse architectures over larger parametrized models, since in all of our experiments pruned networks significantly outperform their larger unpruned counterparts. These results suggest that winning initializations do contain inductive biases that are generic to neural networks, although, as reported by our experiments on the biomedical datasets, their generalization properties can be more limiting than what has so far been observed in the literature.
\end{abstract}

\section{Introduction}
\label{sec:intro}
\begin{figure*}[!htb]
\minipage{0.25\textwidth}
  \includegraphics[width=\linewidth]{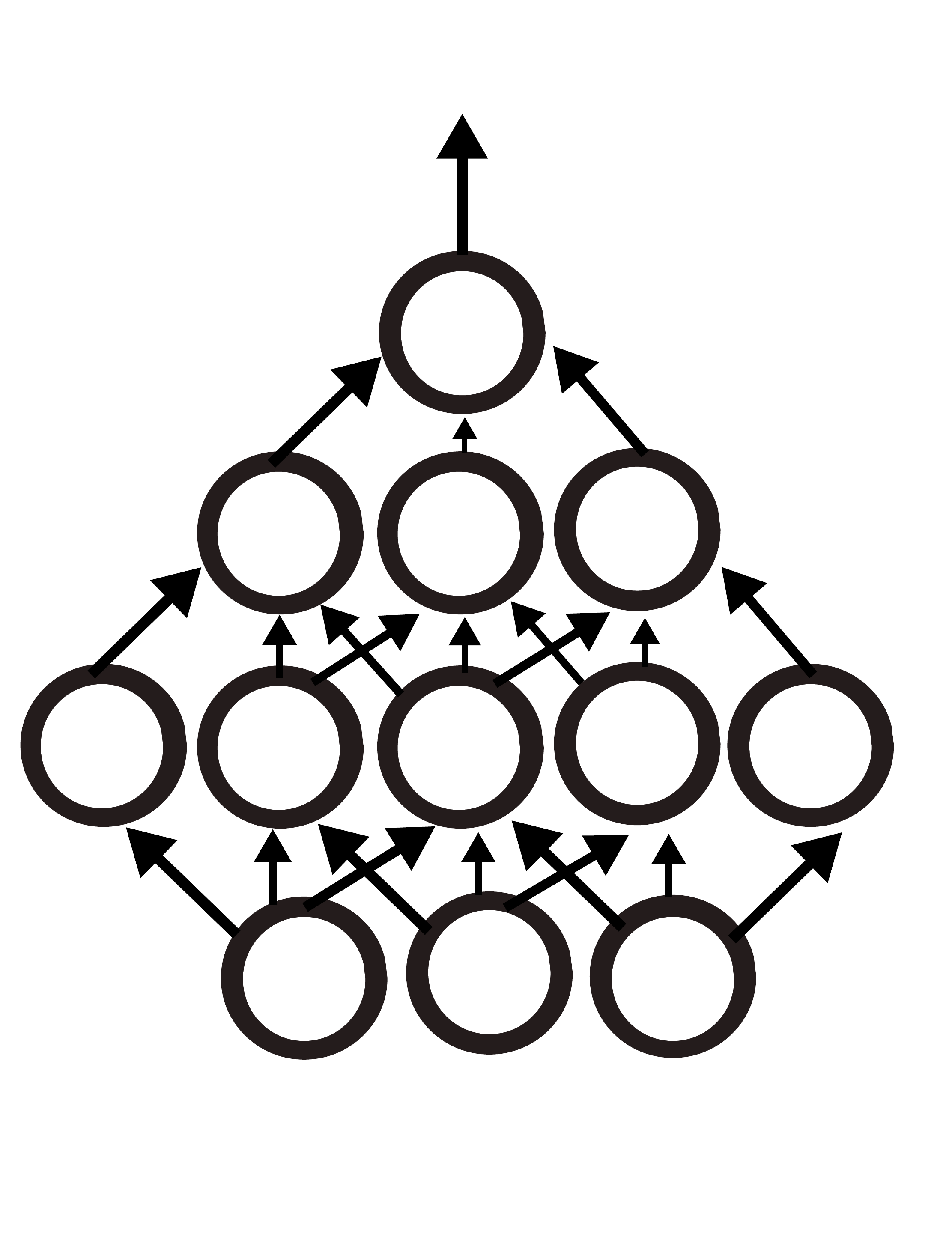}
\endminipage\hfill
\minipage{0.25\textwidth}
  \includegraphics[width=\linewidth]{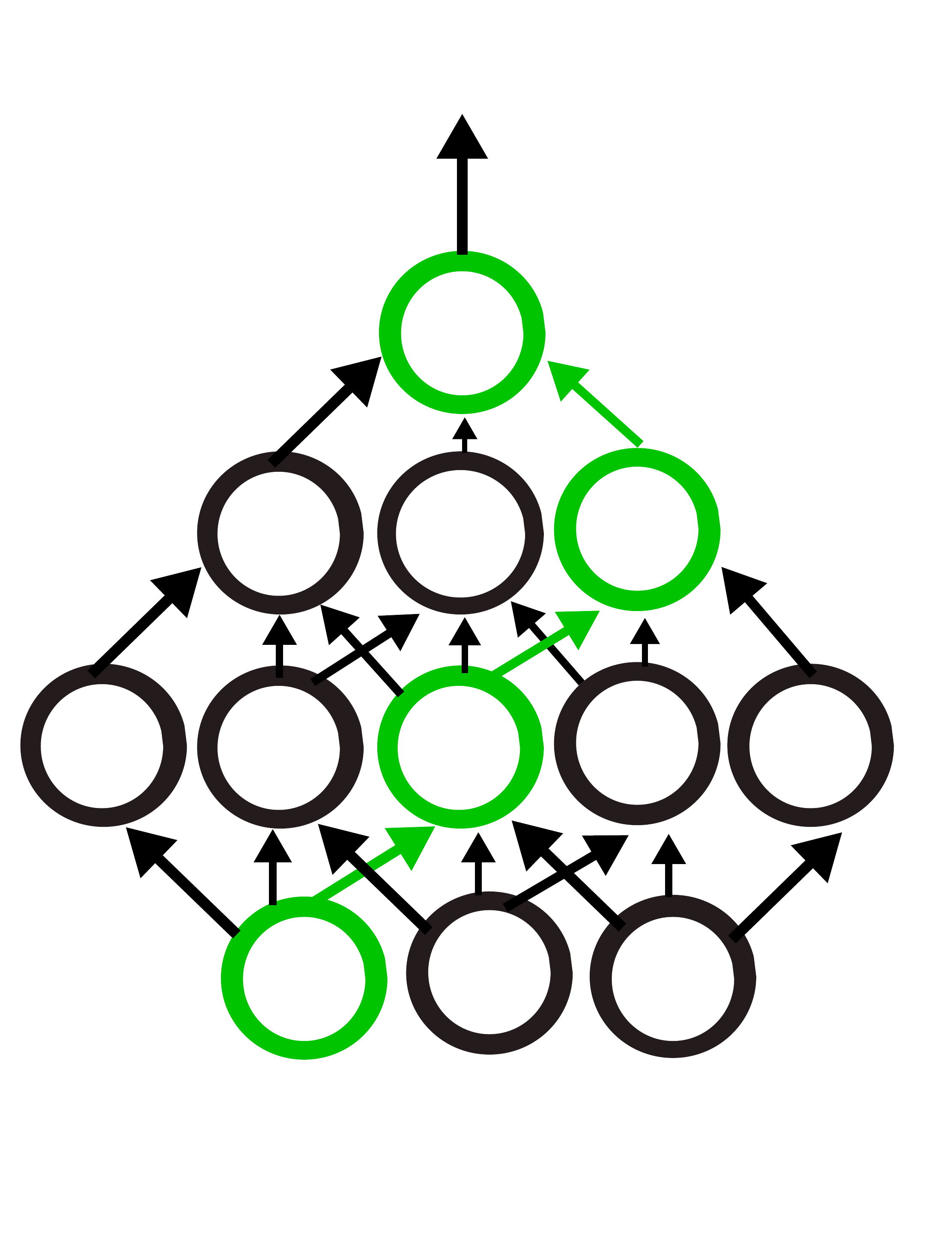}
\endminipage\hfill
\minipage{0.5\textwidth}%
  \includegraphics[width=\linewidth]{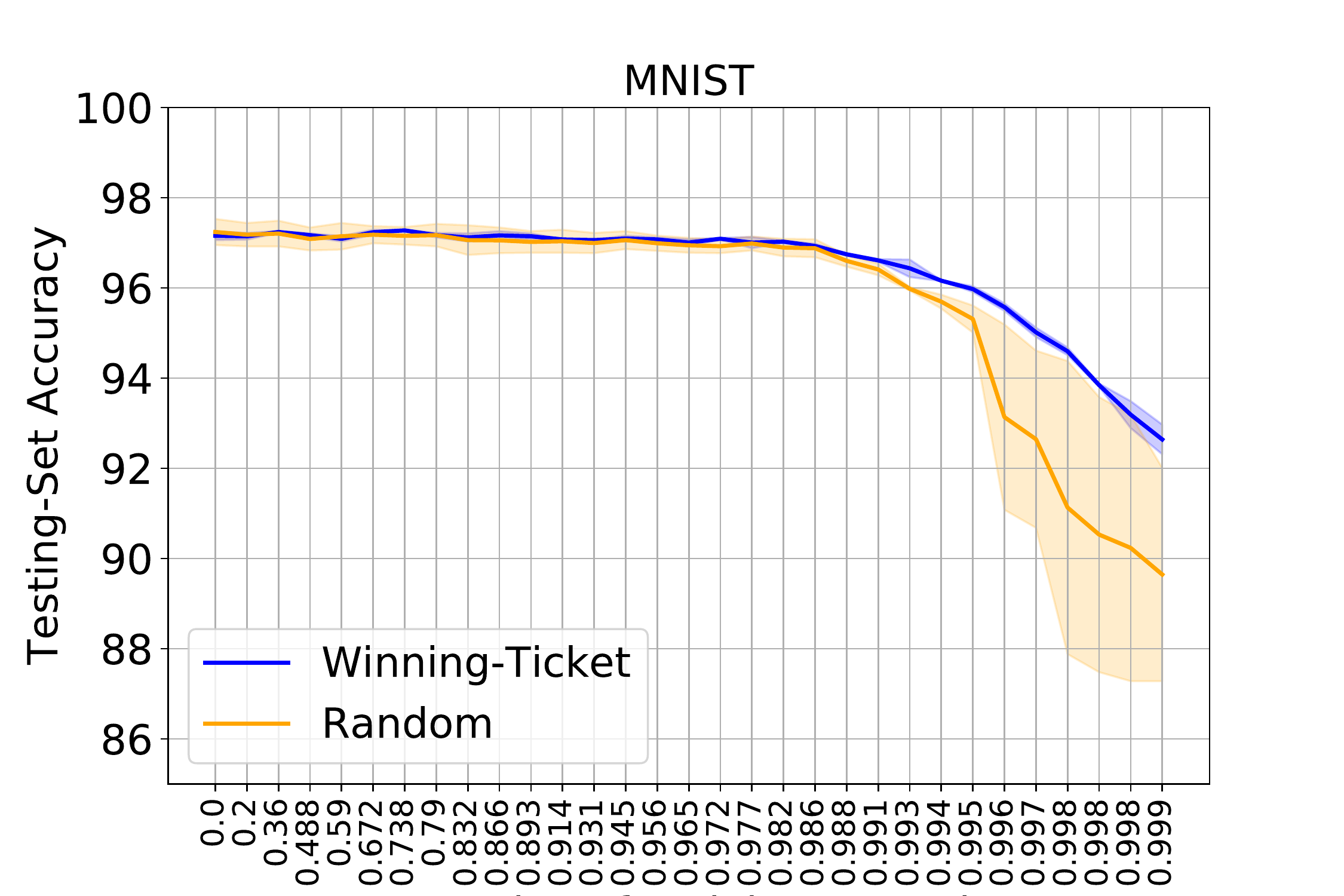}
\endminipage
\caption{A visual representation of the LTH as introduced in \cite{frankle2018lottery}. Let us consider a simplified version of a two hidden layer feedforward neural network as is depicted in the first image on the left. The LTH states that within this neural network there exist multiple smaller networks (represented in green), which perform just as well as their larger counterpart. Training these sparse models from scratch successfully is only possible as long as their weights are initialized with the same values that were also used when the larger (black) model was initialized. As can be seen by the blue curve of the last plot the performance of such pruned models gets barely harmed even when large pruning rates are reached. These models are considered as the winners of the initialization lottery and also perform better than the same models re-initialized randomly (orange line). Results obtained on the MNIST dataset that replicate the findings presented in \cite{frankle2018lottery}.}%
    \label{fig:tickets_visualization}%
\end{figure*}

The ``Lottery-Ticket-Hypothesis" (LTH) \cite{frankle2018lottery} states that within large randomly initialized neural networks there exist smaller sub-networks which, if trained from their initial weights, can perform just as well as the fully trained unpruned network from which they are extracted. This happens to be possible because the weights of these sub-networks seem to be particularly well initialized before training starts, therefore making these smaller architectures suitable for learning (see Fig \ref{fig:tickets_visualization} for an illustration). These sub-networks, i.e., the pruned structure together with their initial weights, are called winning tickets, as they appear to have won the initialization lottery. Since winning tickets only contain a very limited amount of parameters, they yield faster training, inference, and sometimes even better final performance than their larger over-parametrized counterparts \cite{frankle2018lottery,franklestabilizing}. So far, winning tickets are typically identified by an iterative procedure that cycles through several steps of network training and weight pruning, starting from a randomly initialized unpruned network. While simple and intuitive, the resulting algorithm, has unfortunately a high computational cost. Despite the fact that the resulting sparse networks can be trained efficiently and in isolation from their initial weights, the LTH idea has not yet led to more efficient solutions for training a sparse network, than existing pruning algorithms that all also require to first fully train an unpruned network \cite{han2015deep,molchanov2016pruning,dong2017learning,lin2017runtime,zhuang2018discrimination}.

Since the introduction of the idea of the LTH, several research works have focused on understanding what makes some weights so special to be the winners of the initialization lottery. Among the different tested approaches, which will be reviewed in Sec. \ref{sec:related_work}, one research direction in particular has looked into how well winning ticket initializations can be transferred among different training settings (datasets and optimizers), an approach which aims at characterizing the winners of the LTH by studying to what extent their inductive biases are generic \cite{morcos2019one}. The most interesting findings of this study are that winning tickets generalize across datasets, within the natural image domain at least, and that tickets obtained from larger datasets typically generalize better. This opens the door to the transfer of winning tickets between datasets, which makes the high computational cost required to identify them much more acceptable practically, as this cost has to be paid only once and can be shared across datasets.

In this paper, we build on top of this latter work. While Morcos et al. \cite{morcos2019one} focused on the natural image domain, we investigate the possibility of transferring winning tickets obtained from the natural image domain to datasets in non natural image domains. This question has an important practical interest as datasets in non natural image domains are typically scarcer than datasets in natural image domains. They would therefore potentially benefit more from a successful transfer of sparse networks, since the latter can be expected to require less data for training than large over-parametrized networks. Furthermore, besides studying their generalization capabilities, we also focus on another interesting property that characterizes models that win the LTH, and which so far has received less research attention. As originally presented in \cite{frankle2018lottery}, pruned models which are the winners of the LTH can yield a final performance which is better than the one obtained by larger over-parametrized networks. In this work we explore whether it is worth seeking for such pruned models when training data is scarce, a scenario that is well known to constraint the training of deep neural networks. To answer these two questions, we carried out experiments on several datasets from two very different non natural image domains: digital pathology and digital heritage.

\textbf{Research questions and contributions:} this work investigates two research questions. First, we aim at better characterizing the LTH phenomenon by investigating whether lottery winners that are found on datasets of natural images contain inductive biases that are strong enough to allow them to generalize to non-natural image distributions. To do so, we present to the best of our knowledge the first results that study the transferability of winning initializations in this particular training setting. Second, we thoroughly study for the first time whether pruned models that are the winners of the LTH can consistently outperform their larger over-parametrized counterparts in conditions with scarce training data.

\section{Datasets}
\label{sec:datasets}

\begin{figure*}
  \centering
   \includegraphics[width=2.5cm,height=\textheight,keepaspectratio]{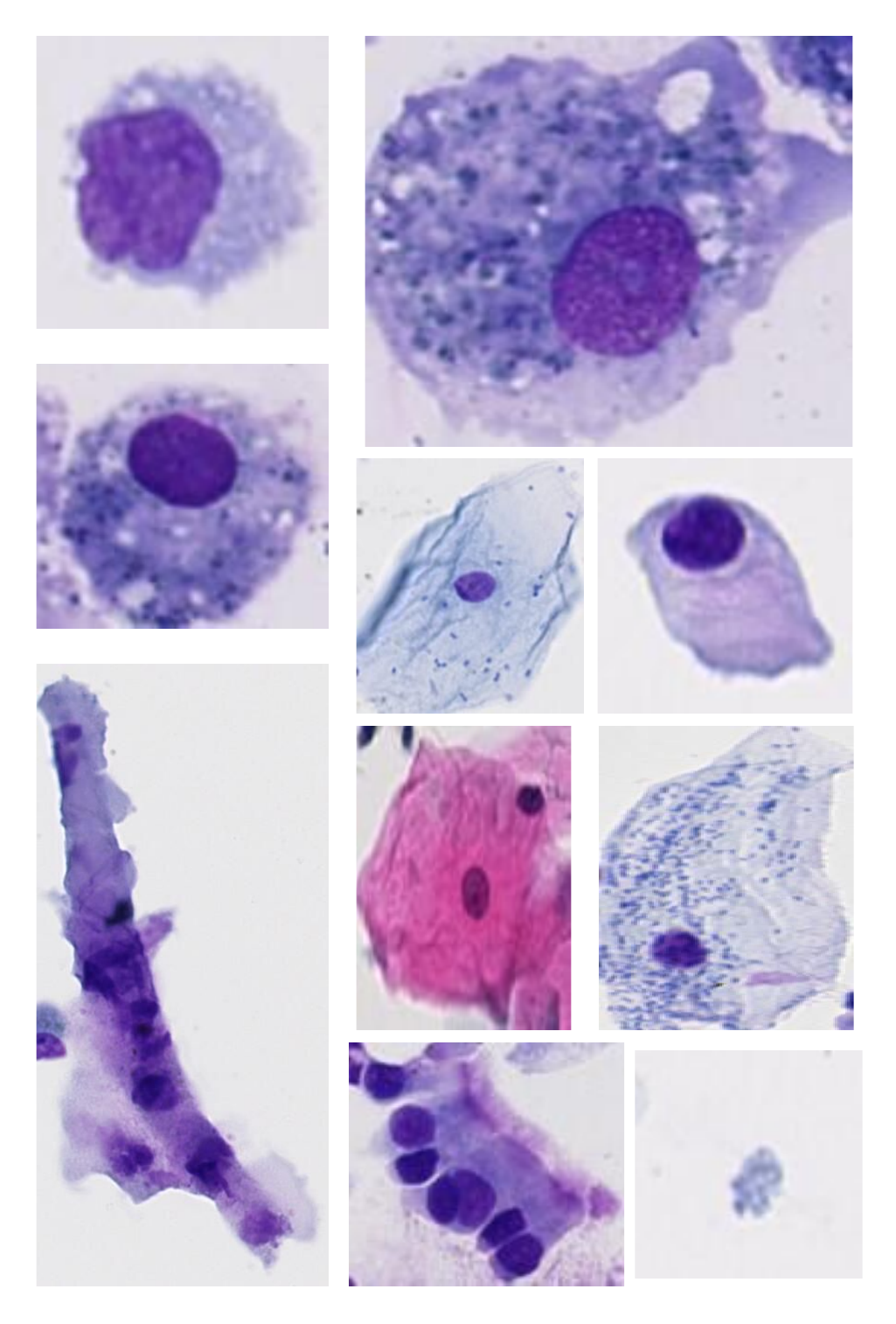}%
  \includegraphics[width=2.5cm,height=\textheight,keepaspectratio]{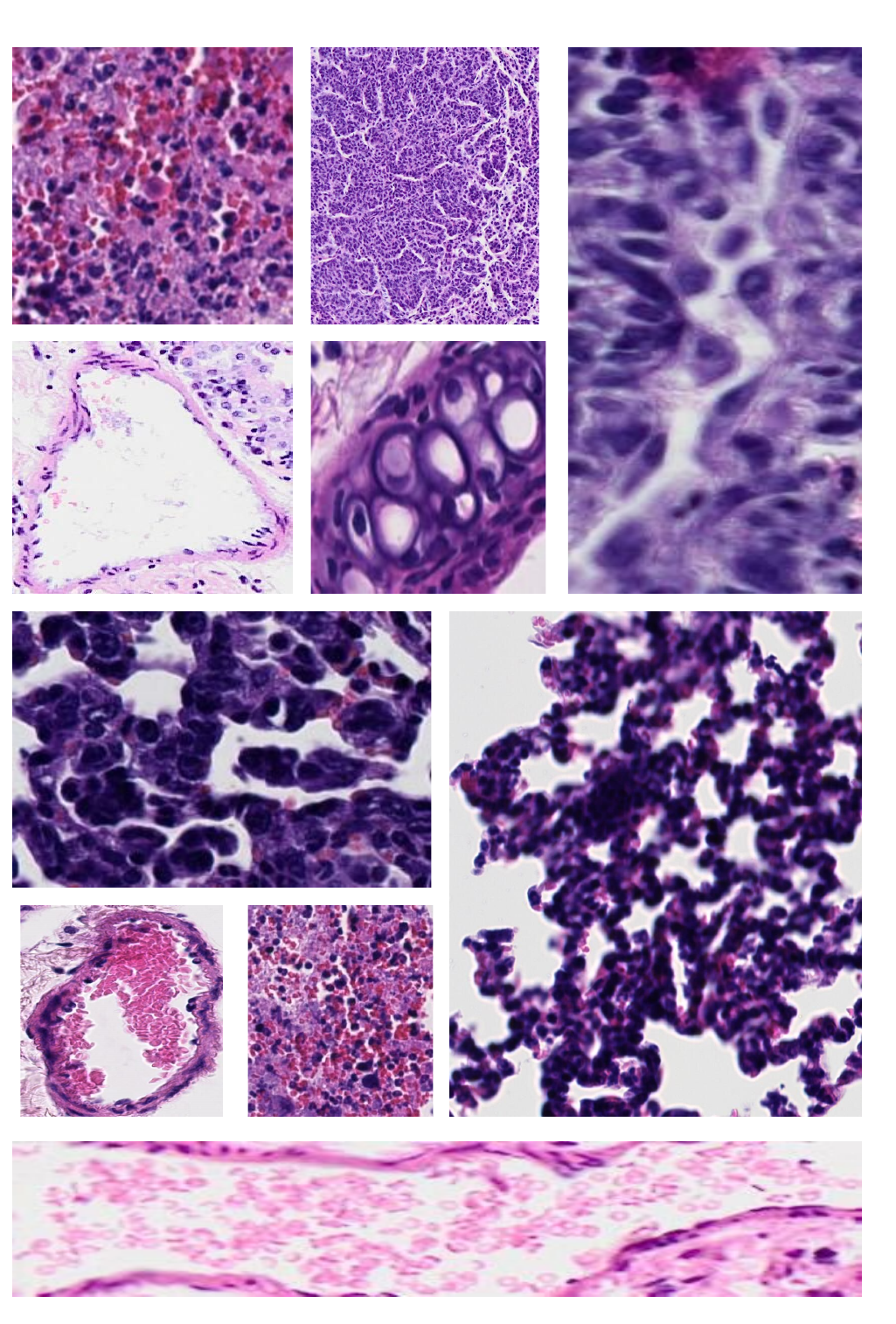}%
  \includegraphics[width=2.5cm,height=\textheight,keepaspectratio]{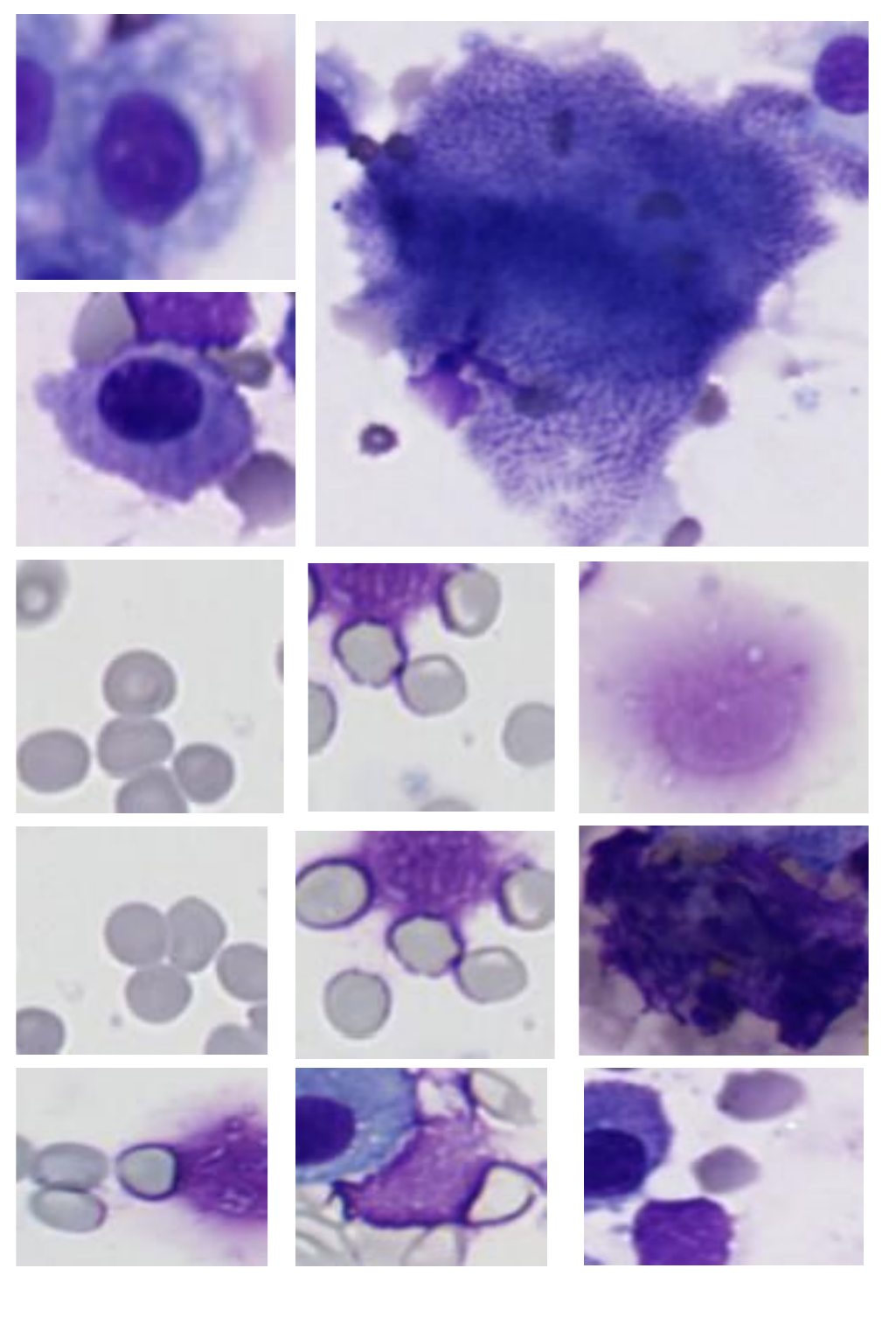}%
    \includegraphics[width=2.5cm,height=\textheight,keepaspectratio]{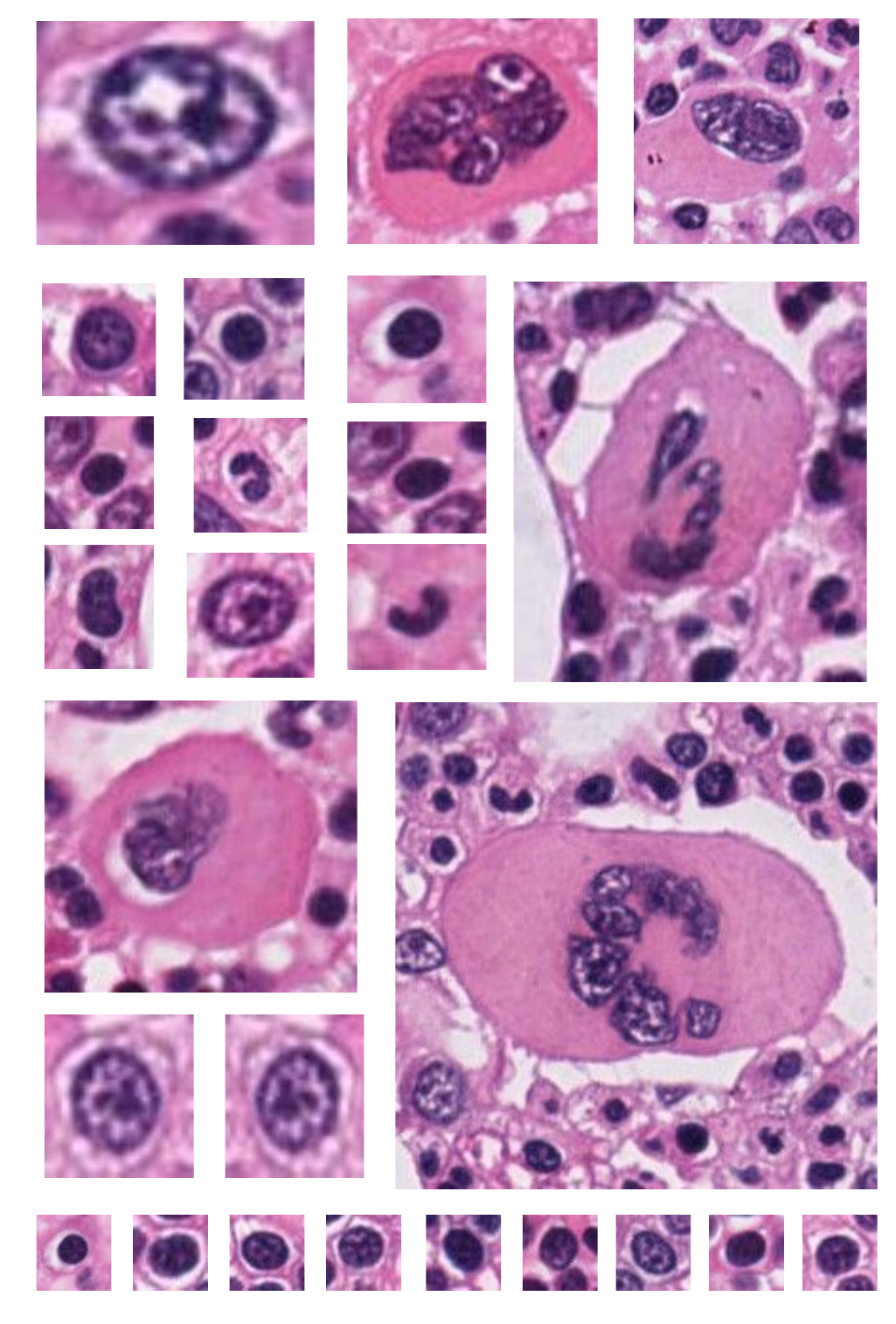}%
  \includegraphics[width=2.5cm,height=\textheight,keepaspectratio]{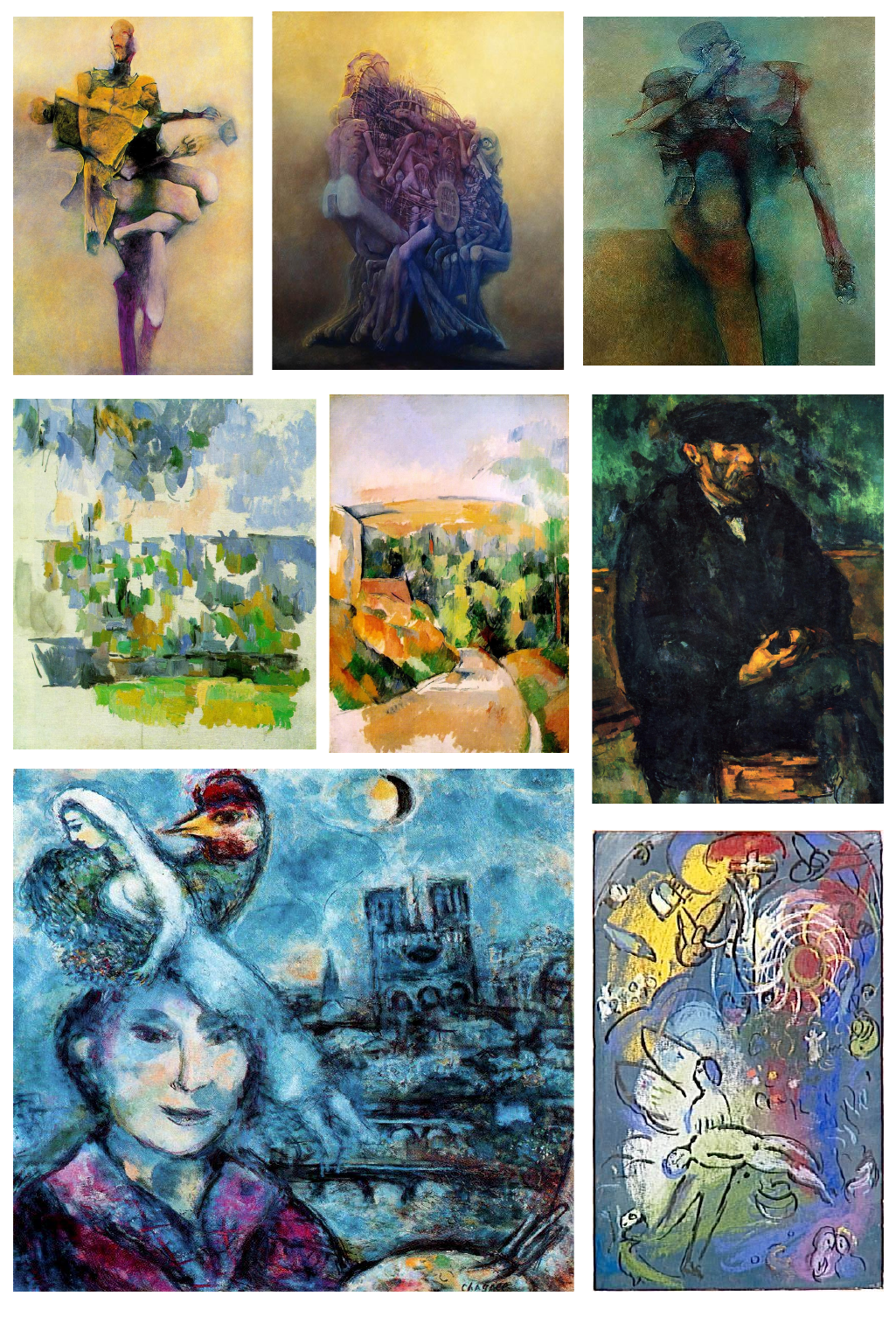}%
  \includegraphics[width=2.5cm,height=\textheight,keepaspectratio]{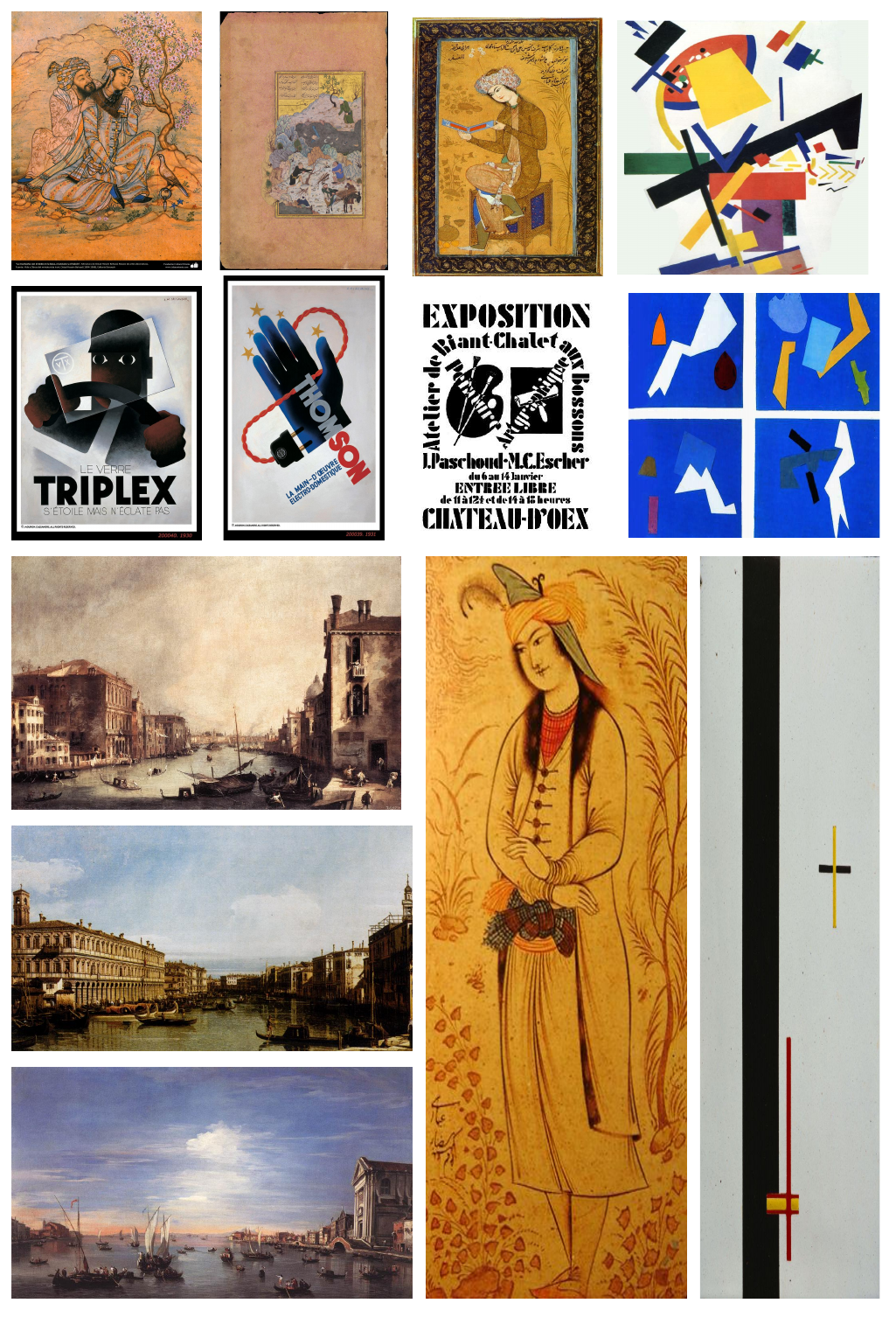}%
\caption{Some image samples that constitute the non-natural image datasets which have been used in this work.
From left to right we have the \texttt{Human-LBA}, \texttt{Lung-Tissues}, \texttt{Mouse-LBA} and \texttt{Bone-Marrow} datasets, while finally we report some examples that represent artworks which come from the field of digital heritage.}
\label{fig:dataset_images}
\end{figure*}

We consider seven datasets that come from two different, unrelated sources: histopathology and digital heritage. Each dataset comes with its training, validation and testing splits. Furthermore the datasets change in terms of size, resolution, and amount of labels that need to be classified.
We report an overview about the size of these datasets in Table \ref{tab:datasets} while a visual representation of the samples constituting these datasets in Fig. \ref{fig:dataset_images}.

The Digital-Pathology (DP) data comes from the Cytomine \cite{maree2016collaborative} web application, an open-source platform that allows interdisciplinary researchers to work with large-scale images. While Cytomine has collected a large number of datasets over the years, in this work we have limited our analysis to a subset of four datasets that all represent tissues and cells from either human or animal organs. Such datasets have already been successfully used in previous work \cite{mormont2018comparison}, that researched whether neural networks pre-trained on natural images could successfully be re-used in the biomedical domain. In this paper, we explore whether an alternative to the transfer-learning approaches presented in \cite{mormont2018comparison} could be based on training pruned networks that are the winners of the LTH. This will allow us to investigate the two research questions introduced in Sec. \ref{sec:intro}: we will explore whether winning initializations that are found on datasets of natural images do generalize to non-natural domains, and whether sparse models winners of the LTH can perform better than larger unpruned models that get trained from scratch. 

Regarding the field of Digital-Humanities (DH) we have created three novel datasets that all revolve around the classification of artworks. We consider two different classification tasks that have already been thoroughly studied by researchers bridging between the fields of Computer Vision (CV) and DH \cite{mensink2014rijksmuseum,strezoski2017omniart,sabatelli2018deep}. The first task consists in identifying the artist of the different artworks, while the second one aims at classifying which kind of artwork is depicted in the different images, a challenge which is usually referred to in the literature as type-classification \cite{mensink2014rijksmuseum,sabatelli2018deep}. When it comes to the artist-classification task we have created two different datasets, which purpose will be better explained in Sec. \ref{sec:additional_studies}. All images are publicly available as part of the WikiArt gallery \cite{phillips2011wiki} and can also be found within the large popular OmniArt dataset \cite{strezoski2018omniart}. Albeit in DH it is actually easier to find large datasets than in histopathology, it is worth mentioning that we have kept the size of these datasets intentionally small in order to fit the research questions introduced in Sec. \ref{sec:intro}. Furthermore, it is also worth noting that there are several additional challenges that need to be overcome when training deep neural networks on artistic collections, which therefore motivate the use of this kind of datasets in this work. The size, texture, and resolution of the images coming from the DH are usually representative of different time periods, artistic movements and might have gone through different digitization processes, which are all reasons that make these datasets largely varied and challenging.

\begin{table*}[ht]
\caption{A brief overview of the seven different datasets which have been used in this work. $N_t$ corresponds to the total amount of samples that are present in the dataset, while $Q_t$ represents the number of classes.}
\centering
\begin{tabular}{lcccccc}
\hline
Dataset & Training-Set & Validation-Set & Testing-Set &$N_t$ & $Q_t$\\
\hline
\texttt{Human-LBA} &4051 &346 &1023 &5420 & 9  \\
\texttt{Lung-Tissues} &4881 &562 &888 &6331 & 10 \\
\texttt{Mouse-LBA} &1722 &716 &1846 &4284 & 8    \\
\texttt{Bone-Marrow} & 522 & 130 & 639 & 1291 & 8\\
\hline
\texttt{Artist-Classification-1} & 3103 & 389 & 389 & 3881 & 20\\
\texttt{Type-Classification} & 2868 & 360 & 360 & 3588 & 20 \\
\texttt{Artist-Classification-2} & 2827 & 353 & 353 & 3533 & 19\\
\hline
\end{tabular}
\label{tab:datasets}
\end{table*}

\section{Experimental Setup}
\label{sec:experimental_setup}
We follow an experimental set-up which is similar to the one that was introduced in \cite{morcos2019one} (and that has been validated by \cite{gohil2019one}).
Let us define a neural network $f(x;\theta)$ that gets randomly initialized with parameters $\theta_0 \sim \mathcal{D_\theta}$ and then trained for $j$ iterations
over an input space $\mathcal{X}$, and an output space $\mathcal{Y}$. 
At the end of training a percentage of the parameters in $\theta_j$ gets pruned, a procedure which results in a mask $m$. The parameters in $\theta_j$ which did not get pruned are then reset to the values they had at $\theta_k$, where $k$ represents an early training iteration. A winning ticket corresponds to the combination between the previously obtained mask, and the parameters $\theta_k$, and is defined as $f(x;m\odot\theta_k)$ \footnote{Note that this formulation generalizes the original version of the LTH \cite{frankle2018lottery} that we have represented in Fig. \ref{fig:tickets_visualization}, where a winning ticket is obtained after resetting the unpruned parameters of the network to the values they had right after initialization, therefore defining a winning ticket as $f(x;m\odot\theta_0)$.}. Constructing a winning ticket with parameters $\theta_k$, instead of $\theta_0$, is a procedure which is known as late-resetting \cite{franklestabilizing}, and is a simple but effective trick that makes it possible to stably find winning initializations in deep convolutional neural networks \cite{franklestabilizing,morcos2019one}. In this study $f(x;\theta)$ comes in the form of a ResNet-50 architecture \cite{han2015deep} which gets trained on the three popular CV natural image datasets CIFAR-10/100 and Fashion-MNIST. Following \cite{han2015deep,morcos2019one}, 31 winning tickets $f(x;m\odot\theta_k)$ of increasing sparsity are obtained from each of these three datasets by repeating 31 iterations of network training and magnitude pruning with a pruning rate of 20\%.
More precisely, at each pruning iteration, the network is trained for several epochs (using early stopping on the validation set as described in the appendix) and then the 20\% of weights with the lowest magnitudes are pruned. The parameters $\theta_k$ that define each of the 31 tickets are then taken as the weights of the corresponding pruned networks at the $k$th epoch of the first pruning iteration. Once these pruned networks are found we aim at investigating whether their parameters $\theta_k$ contain inductive biases that allow them to generalize to the non-natural image domain. To do so we replace the final fully connected layer of each winning ticket with a randomly initialized layer that has as many output nodes as there are classes to classify. We then fine-tune each of these networks on the non-natural image datasets considered in this study.
At the end of training, we study the performance of each winning ticket in two different ways. First, we compare the performance of each network to the performance of a fully unpruned network that gets randomly initialized and trained from scratch. Second, we also compare the performance of winning tickets that have been found on a natural image dataset to 31 new sparse models that are the winners of the LTH on the considered target dataset. Since it is not known to which extent pruned networks that contain weights that are the winners of the LTH on a natural image dataset can generalize to target distributions that do not contain natural images, we report the first results that investigate the potential of a novel transfer-learning scheme which has so far only been studied on datasets from the natural image domain. Moreover, testing the performance of sparse networks that contain winning tickets that are specific to a non-natural image target distribution also allows us to investigate whether it is worth pruning large networks with the hope of finding smaller models that might perform better than a large over-parametrized one. As mentioned in Sec. \ref{sec:intro}, pruned networks that are initialized with the winning weights can sometimes perform better than a fully unpruned network. Identifying such sparse networks leads to a very significant reduction of model size, which can be a very effective way of regularization when training data is scarce.

\section{\uppercase{Results}}
\label{sec:results}
\begin{figure*}[!htb]
\centering
\minipage{0.5\textwidth}
  \includegraphics[width=\linewidth]{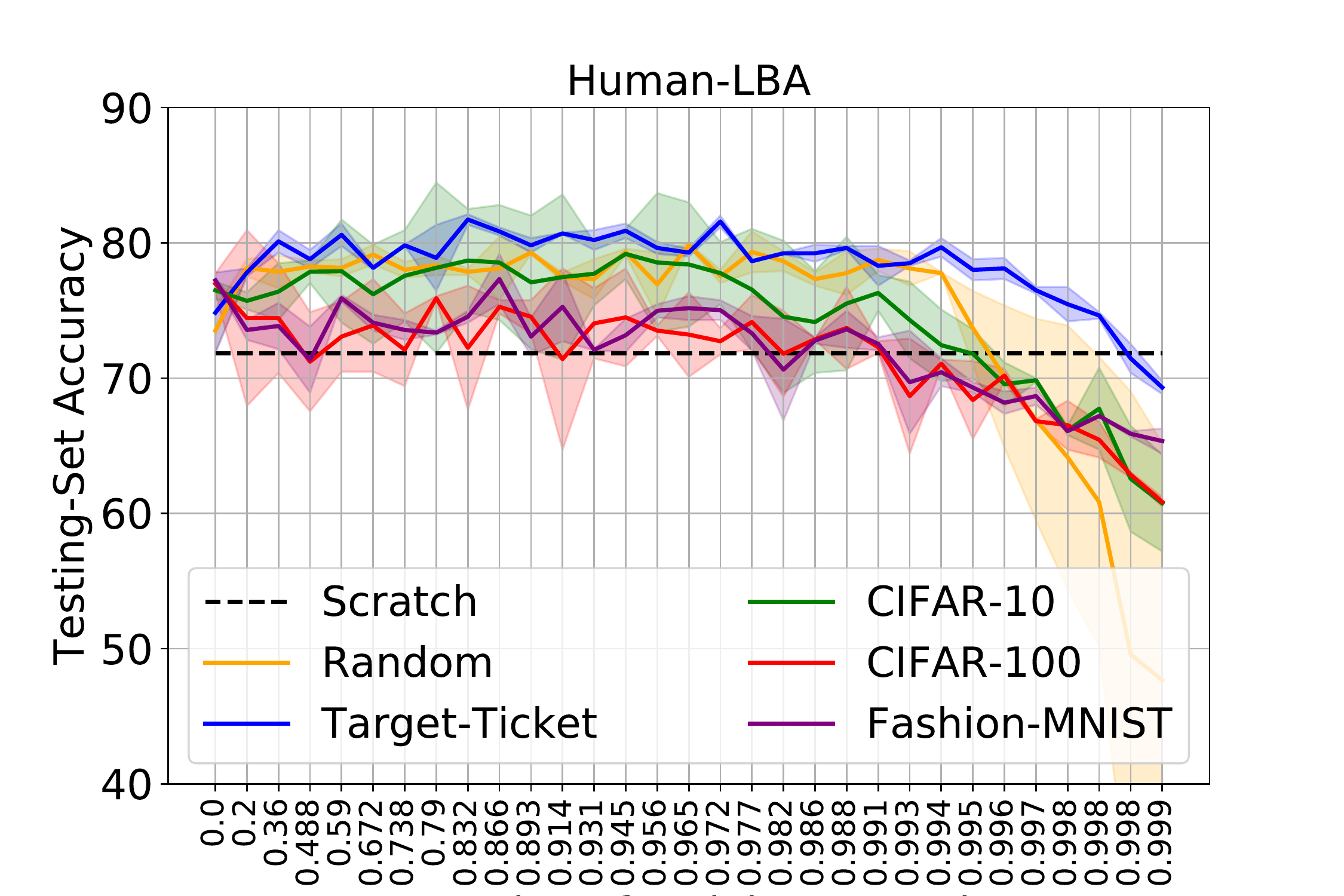}
\endminipage
\minipage{0.5\textwidth}
  \includegraphics[width=\linewidth]{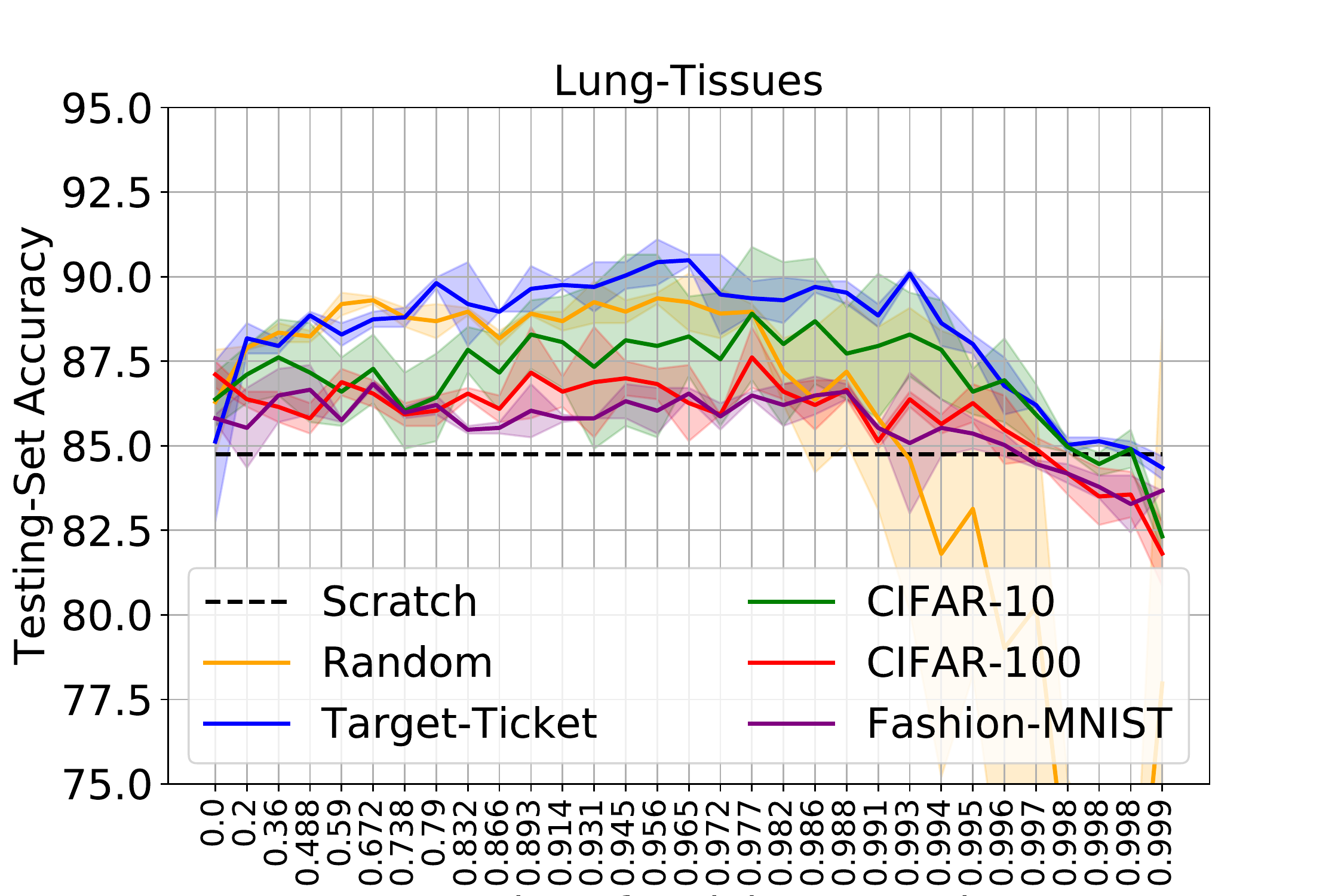}
\endminipage

\minipage{0.5\textwidth}%
  \includegraphics[width=\linewidth]{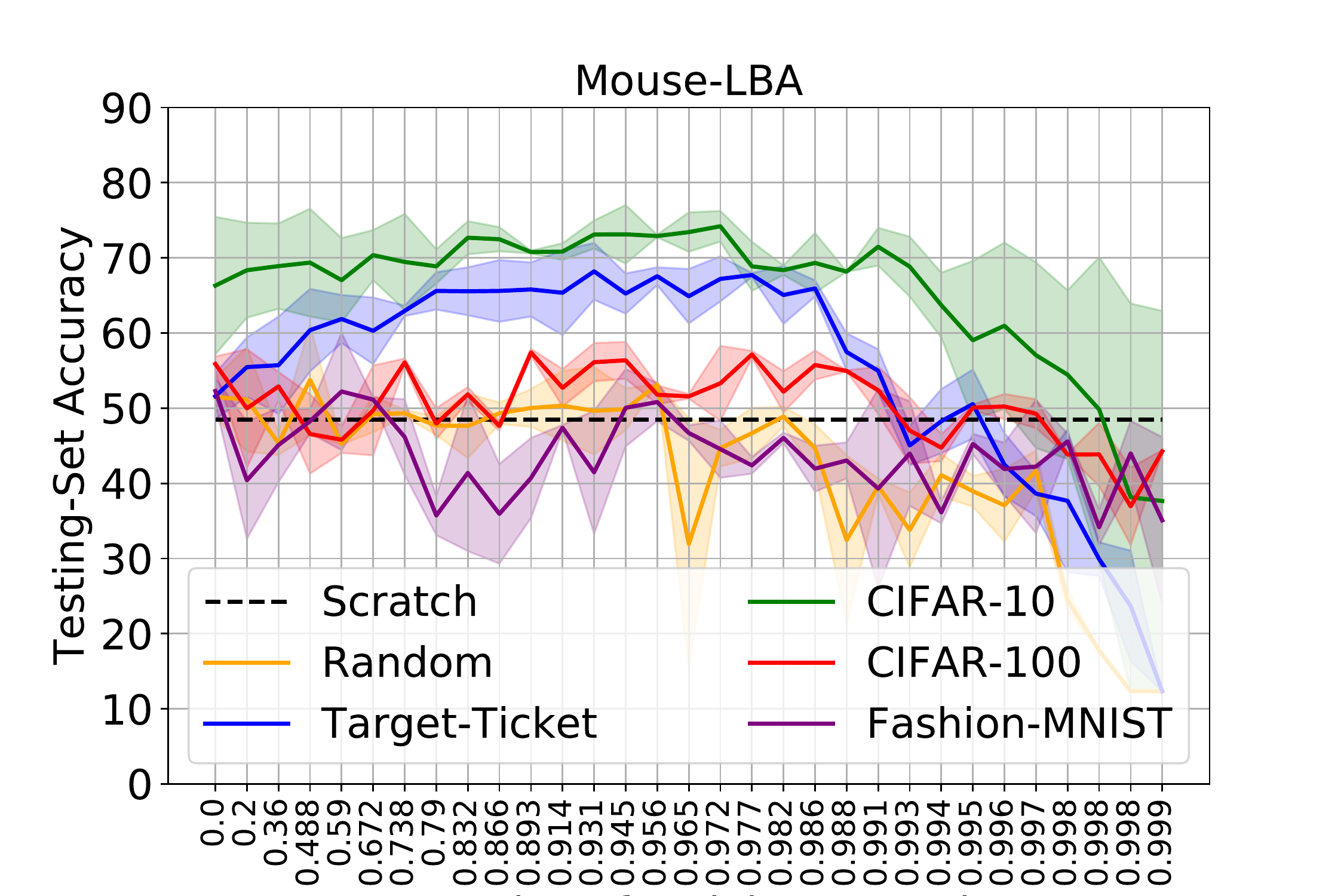}
\endminipage
\minipage{0.5\textwidth}
  \includegraphics[width=\linewidth]{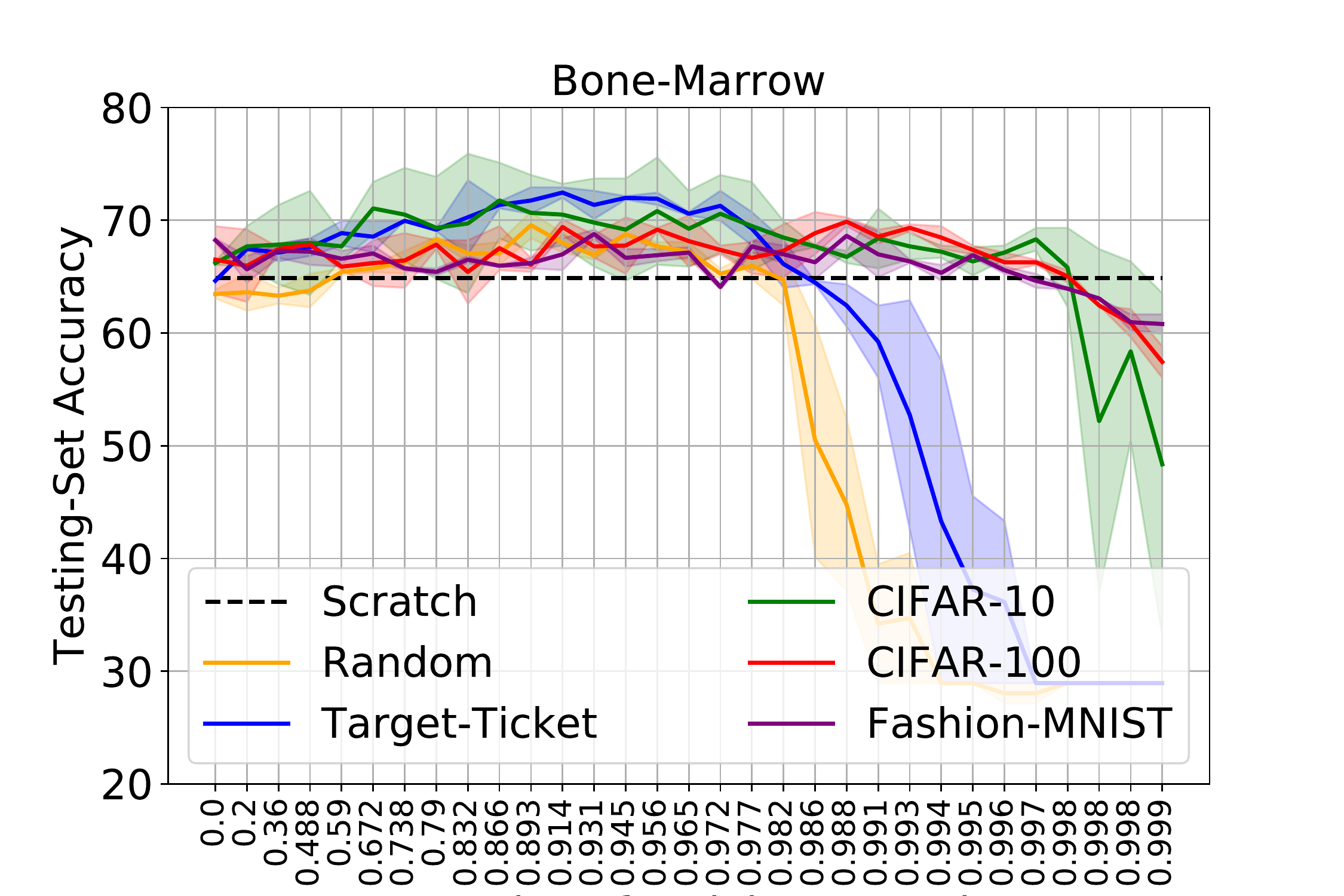}
\endminipage

\minipage{0.5\textwidth}
  \includegraphics[width=\linewidth]{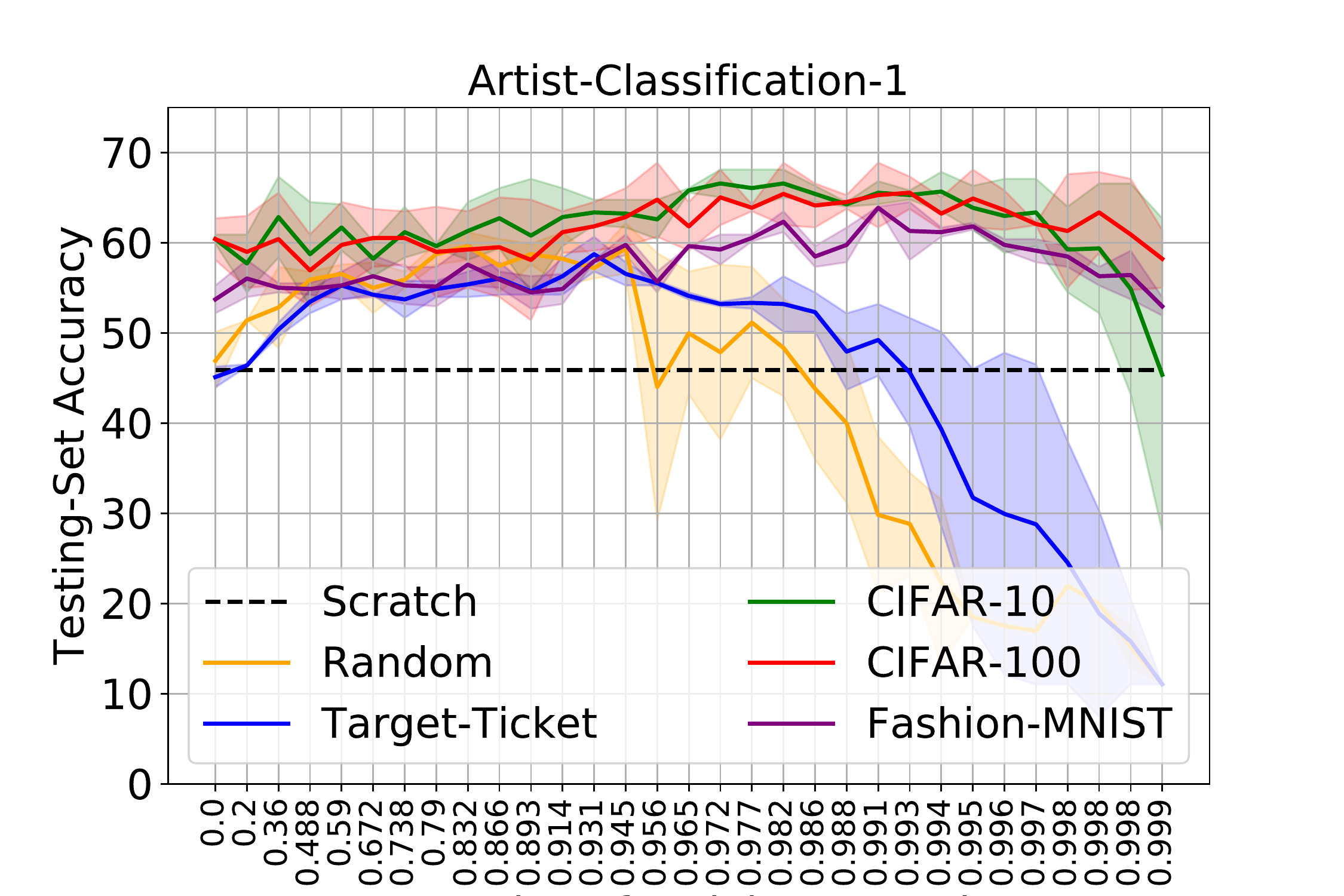}
\endminipage
\minipage{0.5\textwidth}%
  \includegraphics[width=\linewidth]{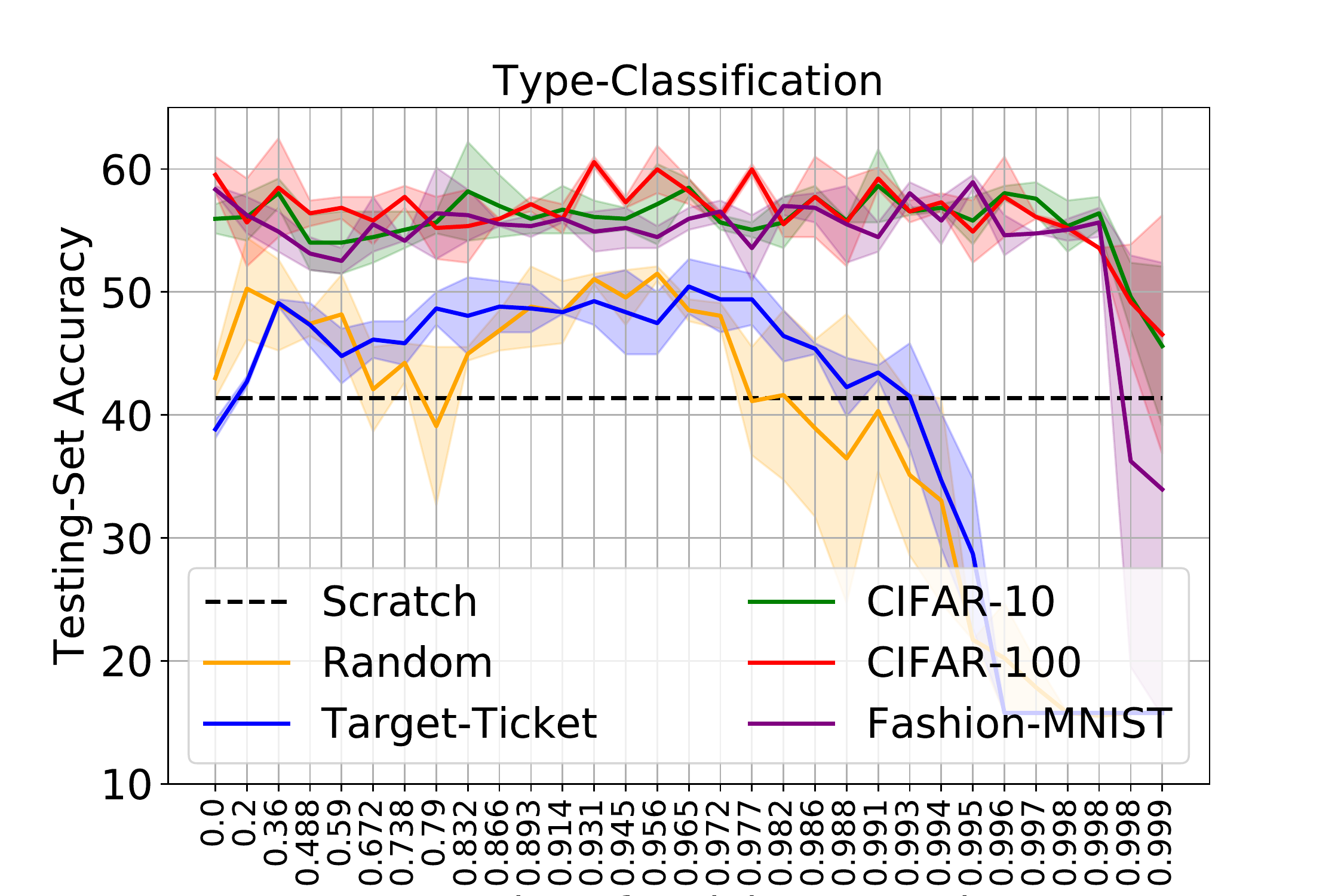}
\endminipage
\caption{An overview of the results showing that sparse models that are the winners of the LTH (represented by the coloured lines) significantly outperform unpruned networks which get randomly initialized and trained from scratch (dashed black line). This happens to be the case on all tested datasets, no matter whether a winning initialization comes from a natural image source or not. It is however worth mentioning that, especially on the biomedical datasets, natural image tickets get outperformed by sparse networks that that are the winners of the LTH on a biomedical dataset. On the other hand this is not the case when it comes to the classification of arts where natural image tickets outperform the ones which are found within artistic collections.}
\label{fig:results}
\end{figure*}

The results of all our experiments are visually reported in the plots of Fig. \ref{fig:results}. Each line plot represents the final performance that is obtained by a pruned model that contains a winning ticket initialization on the final testing-set of our target datasets. This performance is reported on the y-axis of the plots, while on the x-axis we represent the fraction of weights that is pruned from the original ResNet-50 architecture. As explained in the previous section the performance of each winning ticket is compared to the performance that is obtained by an unpruned, over-parametrized architecture that is reported by the black dashed lines. The models that are the winners of the LTH on a natural image dataset are reported by the green, red and purple lines, while the winners of the LTH on a non-natural target dataset are reported by the blue lines. Furthermore, when it comes to the latter lottery tickets, we also report the performance that is obtained by winning tickets that get randomly reinitialized ($f(x;m\odot\theta^{'}_{0})$ with $\theta^{'}_{0} \sim \mathcal{D}_\theta$). These results are reported by the orange lines. Shaded areas around all line plots correspond to $\pm 1$ std. that has been obtained after repeating and averaging the results of our experiments over four different random seeds.

\subsection{On the Importance of Finding Winning Initializations}
We can start by observing that pruned models which happen to be the winners of the LTH either on a natural dataset, or on a non-natural one, can maintain a good final performance until large pruning rates are reached. This is particularly evident on the first three datasets, where models that keep only $\approx 1\%$ of their original weights barely suffer from any drop in performance. This gets a little bit less evident on the last three datasets, where the performance of winning ticket initializations that are directly found on the considered target dataset starts getting harmed once a fraction of $\approx 97$ of original weights are pruned. These results show that an extremely large part of the parameters of a ResNet-50 architecture can be considered as superfluous, therefore confirming the LTH when datasets contain non-natural images. More importantly, we also observe that pruned models winners of the LTH, significantly outperform larger over-parametrized models that get trained from scratch. This can be very clearly seen in all plots where the performance of pruned models is always consistently better than what is reported by the black dashed line. To get a better sense of how much these pruned networks perform better than their larger unpruned counterparts, we report in Table \ref{tab:results} the performance that is obtained by the best performing pruned model, found over all 31 possible pruned models, and compare it to the performance of an unpruned architecture \footnote{The exact fraction of weights which is pruned from an original ResNet-50 architecture is reported in Table \ref{tab:fraction_results}.}. We can observe that no matter which dataset has been used as a source for finding a winning ticket initialization, all pruned networks reach a final accuracy that is significantly higher than the one that is obtained after training an unpruned model from scratch. While in most cases the difference in terms of performance is of $\approx 10\%$ (see e.g. the \texttt{Human-LBA, Lung-Tissues} and the \texttt{Type-Classification} datasets), it is worth highlighting that there are other cases in which this difference is even larger. This is the case for the \texttt{Mouse-LBA} and \texttt{Artist-Classification-1} datasets where a winning ticket coming from the CIFAR-10 dataset performs more than $20\%$ better than a model trained from scratch. These results show that in order to maximize the performance of deep networks it is always worth finding and training pruned models which are the winners of the LTH. 

\begin{table*}[ht!]
\small
\caption{The results comparing the performance that is obtained on the testing-set by the best pruned model winner of the LTH, and an unpruned architecture trained from scratch. The overall best performing model is reported in a green cell, while the second best one in a yellow cell. We can observe that pruned models winners of the LTH perform significantly better than a larger over-parametrized architecture that gets trained from scratch. As can be seen by the results obtained on the \texttt{Mouse-LBA} and \texttt{Artist-Classification-1} datasets the difference in terms of performance can be particularly large ($\approx 20\%$).}
\centering
\begin{tabular}{lcccccc}
\hline
Target-Dataset & Scratch-Training & CIFAR-10 & CIFAR-100 & Fashion-MNIST & Target-Ticket \\
\hline
\texttt{Human-LBA} & $71.85_{\pm 1.12}$ &\cellcolor{yellow!25}$79.17_{\pm1.85}$ &$76.97_{\pm0.73}$ &$77.32_{\pm1.85}$ &\cellcolor{green!25}$81.72_{\pm0.39}$\\
\texttt{Lung-Tissues} &$84.75_{\pm0.81}$ &\cellcolor{yellow!25}$88.90_{\pm1.97}$ &$87.61_{\pm0.90}$ & $87.61_{\pm0.11}$ & $\cellcolor{green!25}90.48_{\pm0.16}$\\
\texttt{Mouse-LBA} &$48.17_{\pm1.18}$ &\cellcolor{green!25}$74.20_{\pm2.04}$ &$57.42_{\pm0.48}$ &$52.27_{\pm1.73}$ &\cellcolor{yellow!25}$68.20_{\pm3.79}$\\
\texttt{Bone-Marrow} &$64.66_{\pm1.36}$ &\cellcolor{yellow!25}$71.75_{\pm3.36}$ &$69.87_{\pm0.39}$&$68.77_{\pm0.39}$&$\cellcolor{green!25}72.55_{\pm0.46}$\\
\hline
\texttt{Artist-Classification-1} & $45.88_{\pm0.42}$ & $\cellcolor{green!25}66.58_{\pm1.54}$ & $\cellcolor{yellow!25}65.55_{\pm1.79}$ & $63.88_{\pm0.12}$ & $58.74_{\pm1.92}$\\
\texttt{Type-Classification} &$41.36_{\pm2.31}$ & $58.63_{\pm2.97}$ & $\cellcolor{green!25}60.56_{\pm0.44}$ & $\cellcolor{yellow!25}58.92_{\pm0.59}$ & $50.44_{\pm2.23}$ \\

\hline
\end{tabular}
\label{tab:results}
\end{table*}

\begin{table*}[ht!]
\small
\caption{Some additional information about the lottery winners which performance is reported in Table \ref{tab:results}. For each winning ticket we report the fraction of weights that is pruned from an original ResNet-50 architecture and that therefore characterizes the level of sparsity of the overall best performing lottery ticket. The results in the Scratch-Training column are not reported since these are unpruned models that are trained from scratch.}
\centering
\begin{tabular}{lccccc }
\hline
Target-Dataset & Scratch-Training & CIFAR-10 & CIFAR-100 & Fashion-MNIST & Target-Ticket \\
\hline
\texttt{Human-LBA} & - & 0.945  & 0.79 & 0.886 & 0.832\\
\texttt{Lung-Tissues} & - & 0.977 & 0.977 & 0.672 &  0.965\\
\texttt{Mouse-LBA} & - & 0.972 & 0.893 & 0.738 & 0.931\\
\texttt{Bone-Marrow} & - & 0.866 & 0.988 & 0.931 & 0.914\\
\hline
\texttt{Artist-Classification-1} & - & 0.972 & 0.993 & 0.991 & 0.931\\
\texttt{Type-Classification} & - & 0.991 & 0.931 & 0.995 & 0.963\\

\hline
\end{tabular}
\label{tab:fraction_results}
\end{table*}

\subsection{On the Generalization Properties of Lottery Winners}
We then investigate whether natural image tickets can generalize to the non-natural setting. Findings differ across datasets. When considering the datasets that come from the field of DP, we can see that, in three out of four cases, winning tickets that are found on a natural image dataset get outperformed by sparse winning networks that come after training a model on the biomedical dataset. This is particularly evident in the results obtained on the \texttt{Human-LBA} and \texttt{Lung-Tissues} datasets where the highest testing-set accuracy is consistently reached by the blue line plots. When it comes to the \texttt{Bone-Marrow} dataset the difference in terms of performance between the best natural image ticket, in this case coming from the CIFAR-10 dataset, and the one coming from the biomedical dataset, is less evident (see Table \ref{tab:results} for the exact accuracies). Furthermore, it is worth highlighting that on the \texttt{Bone-Marrow} dataset, albeit natural image models seem to get outperformed by the ones found on the biomedical dataset, the performance of the latter ones appears to be less stable once extremely large pruning rates are reached.
When it comes to the \texttt{Mouse-LBA} dataset these results slightly differ. In fact, this dataset corresponds to the only case where a natural image source ticket outperforms a non-natural one. As can be seen, by the green line plot, pruned models coming from the CIFAR-10 dataset outperform the ones found on the \texttt{Mouse-LBA} dataset. When focusing our analysis on the classification of arts, we see that the results change greatly from the ones obtained on the biomedical datasets. In this case, all of the natural image lottery winners, no matter the dataset they were originally found on, outperform the same kind of models that were found after training a full network on the artistic collection. We can see from Table \ref{tab:results} that the final testing performance is similar among all of the best natural image tickets. Similarly to what has been noticed on the \texttt{Bone-Marrow} dataset we can again observe that tickets coming from a non-natural data distribution seem to suffer more from large pruning rates. 

These results show both the potential and the limitations that natural image winners of the LTH can offer when they are fine-tuned on datasets of non-natural images. The results obtained on the artistic datasets suggest that winning initializations contain inductive biases that are strong enough to get at least successfully transferred to the artistic domain, therefore confirming some of the claims that were made in \cite{morcos2019one}. However, it also appears that there are stronger limitations to the transferability of winning initializations which were not observed by \cite{morcos2019one}. In fact, our results show that on DP data the best strategy is to find a winning ticket directly on the biomedical dataset, and that winning initializations found on natural image datasets, albeit outperforming a randomly initialized unpruned network, perform worse than pruned models that are the winners of the LTH on a biomedical dataset.

\subsection{Additional Studies}
\label{sec:additional_studies}
To characterize the transferability of winning initializations even more, while at the same time gaining a deeper understanding of the LTH, we have performed a set of three additional experiments which help us characterize this phenomenon even better. 

\subsubsection{Lottery Tickets VS fine-tuned pruned models}
So far we have focused our transfer-learning study on lottery tickets that come in the form of $f(x;m\odot\theta_k)$, where, as mentioned in Sec. \ref{sec:experimental_setup}, $\theta_k$ corresponds to the weights that parametrize a neural network at a very early training iteration. This formalization is however different from more common transfer-learning scenarios where neural networks get transferred with the weights that are obtained at the end of the training process \cite{mormont2018comparison,sabatelli2018deep}. We have therefore studied whether there is a difference in terms of performance between transferring and fine-tuning a lottery ticket with parameters $\theta_k$, and the same kind of pruned network which is initialized with the weights that are obtained once the network is fully trained on a source task. We define these kind of models as $f(x;m\odot\theta_i)$ where $i$ stays for the last training iteration.  We report some examples of this behaviour in the first row of plots presented in Fig. \ref{fig:additional_studies}, where we consider $f(x;m\odot\theta_i)$ models which were trained on the CIFAR-10 and CIFAR-100 datasets, and then transferred and fine-tuned on the \texttt{Human-LBA} dataset. We found that these models overall perform worse than lottery tickets, while also being less robust to pruning. This also shows that on this dataset, the slightly inferior performance of the natural image tickets with respect to the target tickets is not due to the weight re-initialization.

\subsubsection{Transferring tickets from similar non-natural domains}
We investigated whether it is beneficial to fine-tune lottery winners that, instead of coming from a natural image distribution, come from a related non-natural dataset. Specifically we tested whether winning tickets generated on the \texttt{Human-LBA} dataset generalize to the \texttt{Mouse-LBA} one (since both datasets are representative of the field of Live-Blood-Analysis), and whether lottery winners coming from the \texttt{Artist-Classification-1} dataset generalized to the \texttt{Artist-Classification-2} one. We visually represent these results in the central plots presented in Fig. \ref{fig:additional_studies} As one might expect, we found that it is beneficial to transfer winning tickets that come from a related source. Specifically, \texttt{Human-LBA} tickets can perform just as well as winning tickets that are generated on the \texttt{Mouse-LBA} dataset, while at the same time also being more robust to large pruning rates. When it comes to lottery winners found on the \texttt{Artist-Classification-1} dataset we have observed that these tickets can even outperform the ones generated on the \texttt{Artist-Classification-2} one.

\subsubsection{On the size of the training set}
\begin{figure*}[!htb]
\centering
\minipage{0.5\textwidth}
  \includegraphics[width=\linewidth]{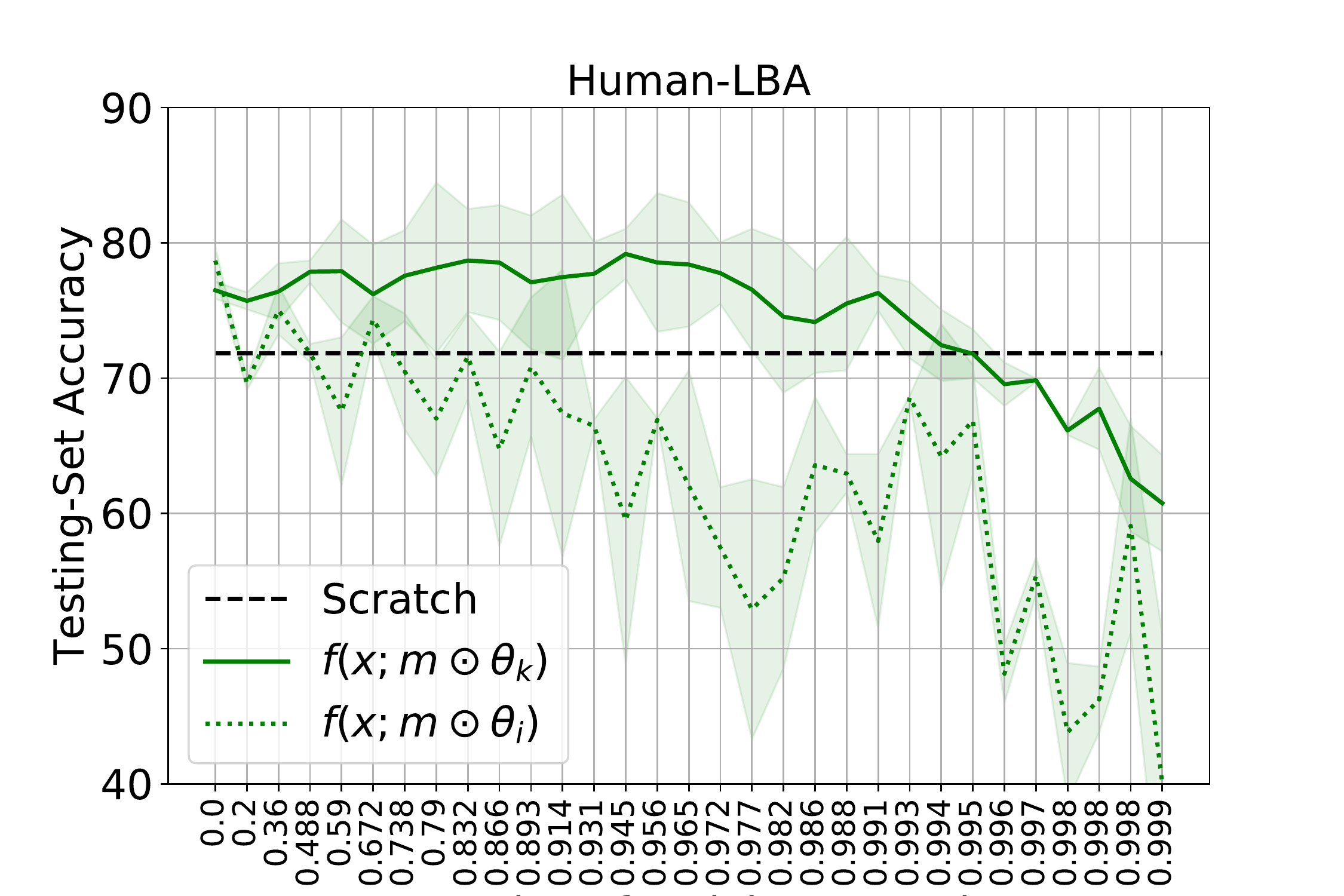}
\endminipage
\minipage{0.5\textwidth}
  \includegraphics[width=\linewidth]{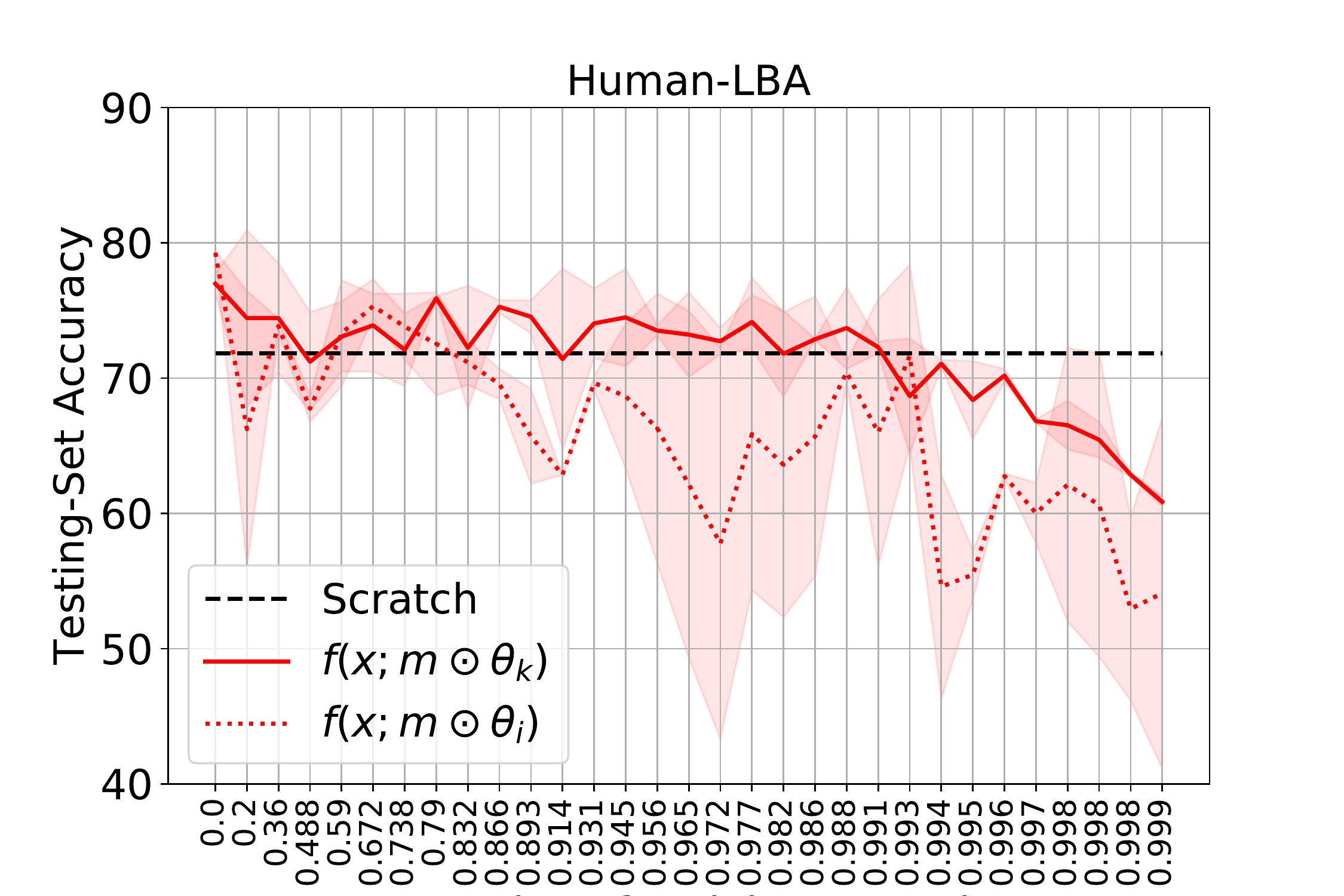}
\endminipage

\minipage{0.5\textwidth}
  \includegraphics[width=\linewidth]{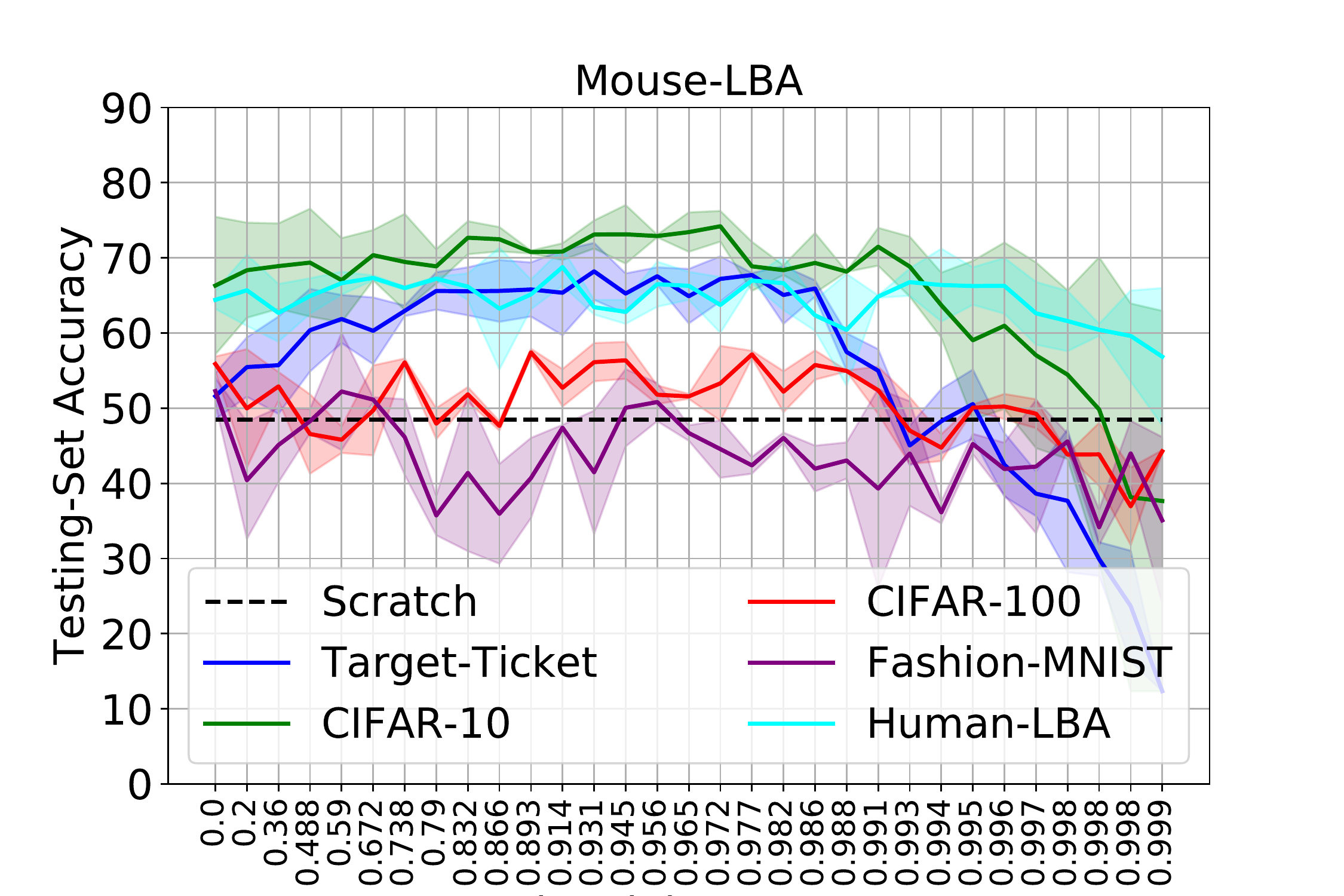}
\endminipage
\minipage{0.5\textwidth}
  \includegraphics[width=\linewidth]{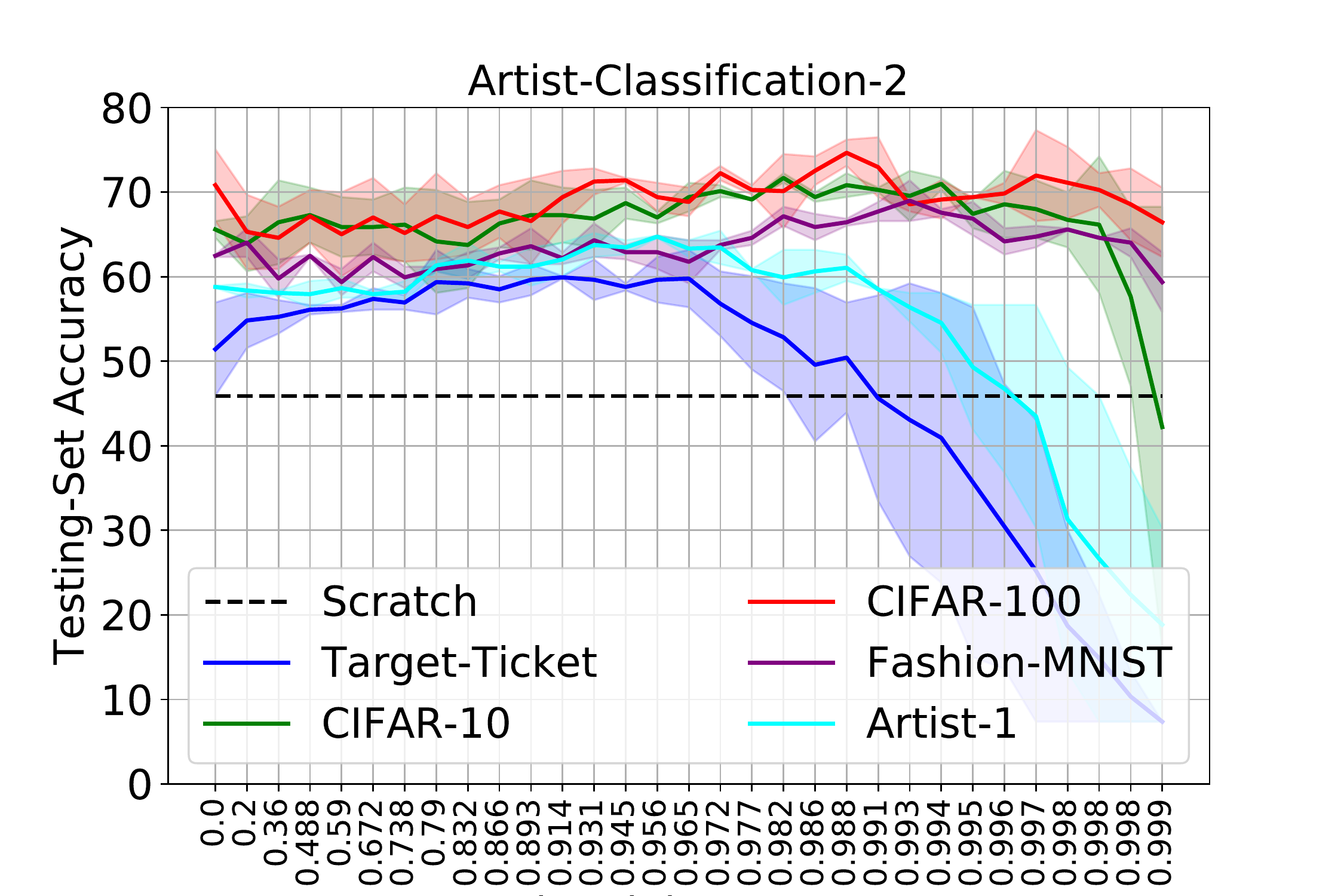}
\endminipage

\minipage{0.5\textwidth}
  \includegraphics[width=\linewidth]{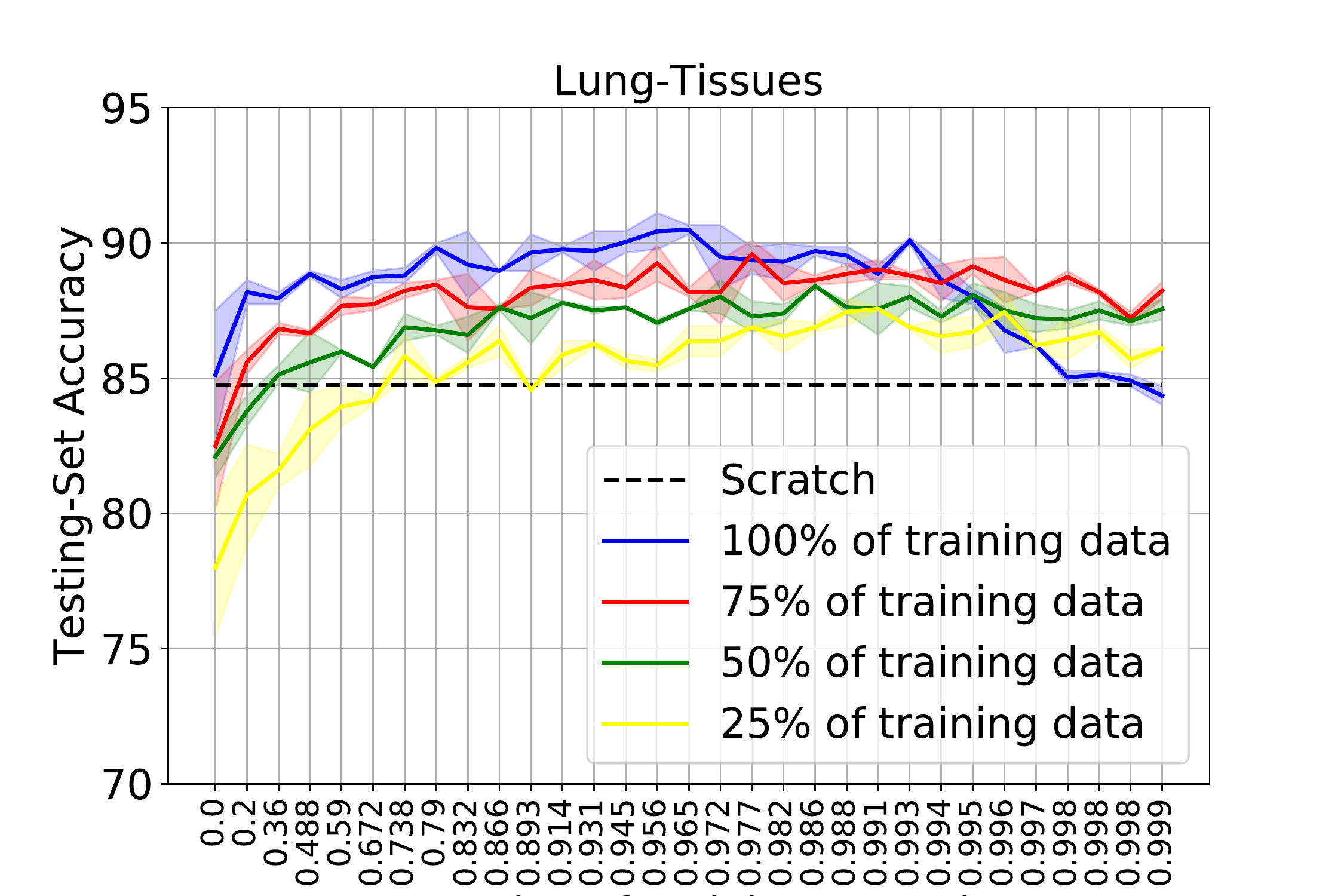}
\endminipage
\minipage{0.5\textwidth}
  \includegraphics[width=\linewidth]{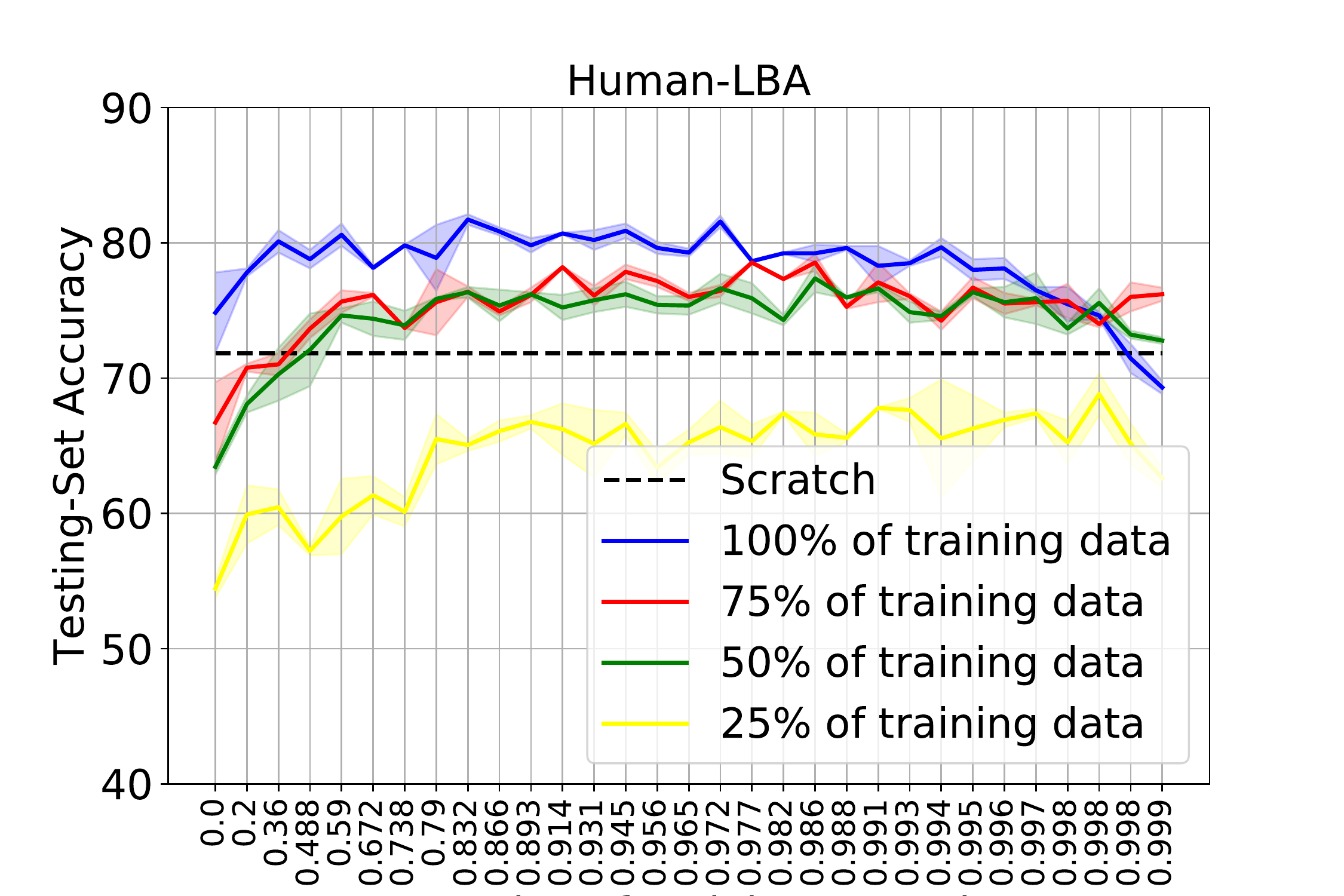}
\endminipage
\caption{A visualization of the results of the additional studies presented in Sec. \ref{sec:additional_studies}. In the first row our study which compares the performance of lottery tickets to the one of fully trained pruned networks. In the second row our results that show some of the benefits that could come from transferring winning tickets generated on similar non-natural distributions. Lastly, in the third row, our study that shows that the stability performance of lottery tickets seems to not be dependent from the size of the training set.
}
\label{fig:additional_studies}
\end{figure*}

We have observed from the blue line plots of Fig. \ref{fig:results} that there are cases in which lottery winners are very robust to extremely large pruning rates (see as an example the first and second plots), while there are other cases in which their performance deteriorates faster with respect to the fraction of weights that get pruned. The most robust performance is obtained by winning tickets that are generated on the \texttt{Human-LBA} and \texttt{Lung-Tissues} datasets, which are the two target datasets that contain the largest amount of training samples. We have therefore studied whether there is a relationship between the size of the training data that is used for finding lottery winners, and the robustness in terms of performance of the resulting pruned models. We generated lottery winners after incrementally reducing the size of the training data by $75\%$, $50\%$ and $25\%$, and then investigated whether we could observe a similar drop in performance as the one which can be seen by the last three blue line-plots of Fig. \ref{fig:results} once a large fraction of weights got pruned. Perhaps surprisingly, we have observed that this was not the case, and as can be seen by the plots represented in the last row of Fig. \ref{fig:additional_studies}, the performance of lottery winners that are found when using only $25\%$ of the training set is just as stable as the one of winning tickets which are generated on the entire dataset. It is however worth mentioning that, albeit the performance of such sparse models is robust, their final performance on the testing set is lower than the one that is obtained by winning tickets that have been trained on the full training data distribution.

\section{Related Work}
\label{sec:related_work}

The research presented in this paper contributes to a better understanding of the LTH by exploring the generalization and transfer-learning properties of lottery tickets. The closest approach to what has been presented in this work is certainly \cite{morcos2019one}, which shows that winning models can generalize across datasets of natural images and across different optimizers. As mentioned in Sec. \ref{sec:experimental_setup}, a large part of our experimental setup is based on this work. Besides the work presented in \cite{morcos2019one}, there have been other attempts that aimed to better understand the LTH after studying it from a transfer-learning perspective. However, just as the study presented in \cite{morcos2019one}, all this research limited its analysis to natural images. In \cite{van2019using} the authors transfer winning tickets among different partitions of the CIFAR-10 dataset, while in \cite{mehta2019sparse} the authors show that sparse models can successfully get transferred from the CIFAR-10 dataset to other object recognition tasks. While these results seem to suggest that lottery tickets contain inductive biases which are strong enough to generalize to different domains, it is worth highlighting that their transfer-learning properties were only studied after considering the CIFAR-10 dataset as a possible source for winning ticket initializations, a limitation which we overcome in this work. It is also worth mentioning that the research presented in this paper is strongly connected to the work presented in \cite{franklestabilizing}. While the first paper that introduced the LTH limited its analysis to relatively simple neural architectures, such as multilayer perceptrons and convolutional networks which were tested on small CV datasets, the presence of winning initializations in larger, more popular convolutional models such as \cite{szegedy2015going} and \cite{he2016deep} trained on large datasets \cite{russakovsky2015imagenet} was only first presented in \cite{franklestabilizing}. Since in this work we have used a ResNet-50 architecture \cite{he2016deep}, we have followed all the recommendations that were introduced in \cite{franklestabilizing}, for successfully identifying the winners of the LTH in larger models. More specifically we mention the late-resetting procedure which resets the weights of a pruned model to the weights that are obtained after $k$ training iterations instead of to the values which were used at the beginning of training (as explained in Sec. \ref{sec:experimental_setup}), a procedure which has shown to be related to \textit{linear mode connectivity} \cite{frankle2019linear}. While the work presented in this paper has limited its analysis to networks that minimize an objective function that is relevant for classification problems, it is worth noting that more recent approaches have identified lottery winners in different training settings. In \cite{yu2019playing} the authors show that winning initializations can be found when neural networks are trained on tasks ranging from natural language processing to reinforcement learning, while in \cite{sun2019learning} the authors successfully identify sparse winning models in a multi-task learning scenario. As future work, we want to study whether lottery tickets can be found on different neural architectures, and also when neural networks are trained on CV tasks other than classification. More specifically we aim at studying whether winners of the LTH, which are found on popular natural image datasets such as \cite{lin2014microsoft} and \cite{everingham2010pascal} when tackling image localization and segmentation tasks, can generalize to non-natural settings which might include the segmentation of biomedical data, or the localization of objects within artworks.       

\section{Conclusion}
\label{sec:conclusion}
We have investigated the transfer learning potential of pruned neural networks that are the winners of the LTH from datasets of natural images to datasets containing non-natural images. We have explored this in training conditions where the size of the training data is relatively small. All of the results presented in this work confirm that it is always beneficial to train a sparse model, winner of the LTH, instead of a larger over-parametrized one. Regarding our study on the transferability of winning tickets we have reported the first results which study this phenomenon under non-natural data distributions by using datasets coming from the fields of digital pathology and heritage. While for the case of artistic data it seems that winning tickets from the natural image domain contain inductive biases which are strong enough to generalize to this specific domain, we have also shown that this approach can present stronger limitations when it comes to biomedical data. This probably stems from the fact that DP images are further away from natural images than artistic ones. We have also shown that lottery tickets perform significantly better than fully trained pruned models, that it is beneficial to transfer lottery winners from different, but related, non-natural sources, and that the performance of lottery tickets is not dependant on the size of the training data. To conclude, we provide a better characterization of the LTH while simultaneously showing that when training data is limited, the performance of deep neural networks can get significantly improved by using lottery winners over larger over-parametrized ones.

\section*{Acknowledgements}
Matthia Sabatelli kindly acknowledges the financial support of BELSPO, Federal Public Planning Service Science Policy, Belgium, in the context of the BRAIN-be project. He also wishes to thank Gilles Louppe for the fruitful brainstorming sessions, and Michela Paganini for the insightful discussions about the Lottery-Ticket Hypothesis and its potential applications to transfer-learning. Lastly, he would like to thank all researchers involved with the Cytomine platform that have worked on collecting and annotating some of the datasets used in this study.

\bibliographystyle{ieee}
{\small
\bibliography{egpaper_final}}

\section*{APPENDIX}

In all of our experiments we have used a ResNet-50 convolutional neural network which has the same structure as the one presented in \cite{han2015deep}.
We have chosen this specific architecture since it has proven to be successful both when used on DP data \cite{mormont2018comparison} as on DH datasets \cite{sabatelli2018deep}. Specifically when it comes to the amount of strides, the sizes of the filters, and the number of output channels, the residual blocks of the network come in the following form:  ($1\times1$, 64, 64, 256) $\times$ 3, (2$\times$2, 128, 128, 512) $\times$ 4, (2$\times$2, 256, 256, 1024) $\times$ 6, (2$\times$2, 512, 512, 2048) $\times$ 3. The last convolution operation of the network is followed by an average pooling layer and a final linear classification layer which has as many output nodes as there is classes to classify in our datasets. Since we only considered classification problems, the model always minimizes the categorical-crossentropy loss function. When feeding the model with the images of the datasets presented in Table \ref{tab:datasets} we extract a random crop of size $224\times224$ and used mini-batches of size 64. No data-augmentation was used. We train the neural network with the Stochastic Gradient Descent (SGD) algorithm with an initial learning rate of $10^{-1}$. SGD is used in combination with Nesterov Momentum $\rho$, set to 0.9, and a weight decay factor $\alpha$ set to $10^{-5}$. Training is controlled by the early-stopping regularization method which stops the training process as soon as the validation loss does not decrease for five epochs in a row. When it comes to the parameters used for pruning we follow a magnitude pruning scheme as the one presented in \cite{han2015learning} which has a pruning-rate of $20\%$. In order to construct winning-tickets we have used the late-resetting procedure with $k=2$. We summarize all this information in Table \ref{tab:hyperparameters}. 

\begin{table}[H]
\centering
\begin{tabular}{l | c |  }
Hyperparameter \\
\hline \hline
Neural Network & ResNet-50 \\ 
Weight-Initialization & Xavier \\
Optimizer & SGD \\
Size of the mini-batches & 64 \\ 
Learning-rate & $10^{-1}$ \\
Momentum $\rho$ & 0.9 \\
Decay-Factor $\alpha$ & $10^{-5}$ \\
Annealing-epochs & $[50,60,75]$\\
Early-Stopping & 5 \\ 
Pruning-Rate & 0.20 \\
Late-resetting $k$ & 2 \\ 
\end{tabular}
\caption{Hyperparameters for our experimental setup.}
\label{tab:hyperparameters}
\end{table}

\end{document}